\documentclass[10pt,journal,compsoc]{IEEEtran}

\ifCLASSOPTIONcompsoc
  \usepackage[nocompress]{cite}
\else
  \usepackage{cite}
\fi
\ifCLASSINFOpdf
\else
\fi
\usepackage{mathrsfs}
\usepackage{amsbsy}
\usepackage{amssymb}
\usepackage{amsmath}
\usepackage{tikz}
\usepackage{multirow}
\usepackage{float}
\usepackage{setspace}
\usepackage{graphicx}
\usepackage{subfigure}
\usepackage{algorithmic}

\usepackage{mdwmath}
\usepackage{mdwtab}
\usepackage{booktabs}
\begin{document}
%
\title{ADMM-Net: A Deep Learning Approach for Compressive Sensing MRI}
%
%
%
%

\author{Yan Yang,
        Jian Sun$^{\ast}$,
        Huibin Li,
        and Zongben Xu

\IEEEcompsocitemizethanks{\IEEEcompsocthanksitem The authors are with the Institute for Information
and System Sciences, School of Mathematics and Statistics,
Xi`an Jiaotong University, Xi`an, 710049, China.\protect\\
E-mail: yangyan92@stu.xjtu.edu.cn, \{jiansun,huibinli,zbxu\}@mail.xjtu.

edu.cn.}
\thanks{Manuscript received xxx, xxx; revised xxx, xxx.}}

\IEEEtitleabstractindextext{%
\begin{abstract}
Compressive sensing (CS) is an effective approach for fast Magnetic Resonance Imaging (MRI). It aims at reconstructing MR images from a small number of under-sampled data in $k$-space, and accelerating the data acquisition in MRI. To improve the current MRI system in reconstruction accuracy and speed, in this paper, we propose two novel deep architectures, dubbed ADMM-Nets in basic and generalized versions. ADMM-Nets are defined over data flow graphs, which are derived from the iterative procedures in Alternating Direction Method of Multipliers (ADMM) algorithm for optimizing a general CS-based MRI model. They take the sampled $k$-space data as  inputs and output reconstructed MR images.
Moreover, we extend our network to cope with complex-valued MR images.
In the training phase, all parameters of the nets, e.g., transforms, shrinkage functions, etc., are discriminatively trained end-to-end.
In the testing phase, they have computational overhead similar to ADMM algorithm  but use optimized parameters learned from the data for CS-based reconstruction task.
We investigate different  configurations in network structures and conduct extensive experiments on MR image reconstruction under different sampling rates.
Due to the combination of the advantages in model-based approach and deep learning approach,
the ADMM-Nets achieve  state-of-the-art reconstruction accuracies with fast computational speed.
\end{abstract}

}

\maketitle

\IEEEdisplaynontitleabstractindextext

%
\IEEEpeerreviewmaketitle

\ifCLASSOPTIONcompsoc
\IEEEraisesectionheading{\section{Introduction}\label{sec:introduction}}
\else
\section{Introduction}
\label{sec:introduction}
\fi

%
%
%
%
\IEEEPARstart{C}{ompressive} sensing~(CS), which aims at recovering a signal  allowing for
data sampling rate much lower than Nyquist rate, is a popular approach in  the fields of  signal processing
and machine learning~\cite{donoho2006}. Nowadays, the CS method has been introduced for  magnetic resonance imaging (MRI), leading to one of the most successful CS application, which is called compressive sensing MRI~(CS-MRI)~\cite{lustig2008}.
MRI is a non-invasive  and widely used imaging technique providing both functional and anatomical information for clinical diagnosis,
including  data acquisition and image reconstruction processing.
Imaging quality greatly affects the subsequent image analysis  and processing which are useful for both doctors and computers.
However, long scanning and waiting time may result in motion artifacts and patients' disamenity.
All these facts imply that imaging speed for MRI is a fundamental challenge.
CS-MRI methods handle this problem by
reconstructing high quality MR images
from a few sampling data in $k$-space (i.e., Fourier space), which reduces the scanning time significantly.

Generally, CS-MRI is formulated as a penalized inverse problem, and the solution to the problem is taken as the reconstructed MR image~\cite{lustig2008,lustig2007sparse}.
Regularization of the model related to the data prior 
is a key component in a CS-MRI model to improve imaging precision. According to the CS theory, signal sparsity is an important prior to remove the aliasing artifacts due to undersampling in $k$-space~\cite{lustig2007sparse}. In this way, sparse regularization can be explored in specific transform domain~\cite{lustig2007sparse,Usman2011k} or
general dictionary-based subspace~\cite{ravishankar2011mr} to either achieve a higher acceleration factor or to improve the reconstruction performance.
The corresponding sparse regularization is usually defined by $l_{q}$  ($q \in [0, 1]$) regularizer.
In addition to the sparse representation of the image, nonlocal similarity property of images is also widely utilized in CS-MRI models~\cite{Liang2011Sensitivity,qu2014magnetic}.
To improve the reconstruction quality, the combination of both  local and nonlocal information is investigated in~\cite{JCTV2017}.
All of these approaches determine the formulation of sparse representation and sparse regularization by personal experiences, which is usually sub-optimal considering  the requirements of quality and speed for  MRI reconstruction.
In summary, it is still a challenging task in CS-MRI to choose  optimal  transform domain / subspace and the corresponding regularization.

To optimize a CS-MRI model, there are three types of algorithms including gradient-based algorithm~\cite{lustig2007sparse}, variable splitting algorithm~\cite{Yin2008Bregman,Li2013An,yang2010fast} and operator splitting algorithm~\cite{Ma2008An}.
Alternating Direction Method of Multipliers (ADMM) is a widely utilized  variable splitting algorithm in CS-MRI, which has been proven to be efficient and generally applicable with convergence guarantee~\cite{boyd2011distributed,Wang2015Global}.
It considers the augmented Lagrangian function of a given CS-MRI model, and splits variables into subgroups, which can be alternatively optimized by solving a few simple subproblems. Although ADMM is generally efficient for optimization, it is also not trivial to determine the optimal parameters (e.g., update rates, penalty parameters) influencing the reconstruction accuracy and speed in CS-MRI.

Recently, deep neural networks owing to strong learning ability from data have achieved exciting successes for image classification and segmentation~\cite{Krizhevsky2012ImageNet,Schmidhuber2014Deep}. Moreover, regression-type deep networks provide
state-of-the-art performance in image denoising and super-resolution as well~\cite{Zhang2017Beyond,Kim2016Accurate}.
In this work, we are interested in bridging the deep learning approach and inverse problem of compressive sensing  with application in  MR image reconstruction.

In this paper, we design two effective deep architectures inspired by ADMM algorithm optimizing a CS-MRI model to reconstruct high-quality MR images from under-sampled $k$-space data.
We first  define a deep architecture represented by a data flow graph, which derived from the ADMM iterative procedures for optimizing a general CS-MRI model.
The operations in ADMM  are represented as graph nodes, and the data flow between two operations is represented by a directed edge.
Then we generalize this data flow graph  to two different deep networks, dubbed {\bf\emph{Basic-ADMM-Net}} and {\bf\emph{Generic-ADMM-Net}}
inspired by two different versions of ADMM algorithm.
Furthermore, we extend our network to  cope with the complex-valued MR image, dubbed {\bf\emph{Complex-ADMM-Net}}.
These deep networks  consist of multiple stages, each of which corresponds to an iteration in ADMM algorithm.
Given an under-sampled $k$-space data, it flows over the network and outputs a reconstructed MR image.
All the parameters (e.g., image transforms, shrinkage functions, penalty parameters, update rates, etc.) in the deep networks can be discriminatively learned from training pairs of under-sampled data in $k$-space and reconstructed image using fully sampled data by
L-BFGS optimization and back-propagation~\cite{L1998Gradient} over the deep architectures.
All the experiments demonstrate that the proposed deep networks are effective both in reconstruction accuracy
and speed.

The main contributions of this study can be summarized as follows.
\begin{itemize}
\item We propose two novel deep ADMM-Nets by reformulating the ADMM algorithms solving a general CS-MRI model to deep networks for CS-MRI.
 The parameters in the CS-MRI model and the ADMM algorithm are all discriminatively learned  from data.

  \item Extensive experiments show that the ADMM-Nets achieve state-of-the-art accuracy in MR image reconstruction with fast computational speed.
 \item  Our proposed ADMM-Nets naturally combine the merits of traditional CS-MRI model and deep learning approach, which can be potentially applied to other inverse problems, such as image deconvolution, general compressive sensing applications.
\end{itemize}


The preliminary version of this work has been presented earlier in a conference~\cite{Yang2016admm1}.
This paper extends the initial version from several aspects to advance our approach.
First, we generalize the ADMM-Net to a more general network structure (i.e., Generic-ADMM-Net) achieving  higher MR image reconstruction quality.
Second, we extend the network to reconstruct the complex-valued MR image which is more useful in clinical diagnosis.
Third, we extensively evaluate the ADMM-Nets with different  widths and depths, and demonstrate the superiorities of the networks by more comparative experiments.
We also compare with several recently published deep learning methods in compressive sensing MRI and
confirm that our method is  advantageous among these deep learning methods.

\section{Related work}
\subsection{Model-based CS-MRI Methods}
Compressive sensing has shown considerable promise in
accelerating the speed of MR acquisition through sparse sampling in $k$-space, which has been applied to various MRI applications such as, multi-contrast MRI~\cite{Huang2012Fast}, dynamic  MRI~\cite{Usman2011k,Zhao2012Image}, etc.
Many efforts have been made in model-based CS-MRI method to apply
sparse regularization in the gradient domain\cite{block2007undersampled},
wavelet transform domain~\cite{qu2012undersampled}, discrete cosine transform domain~\cite{lustig2007sparse} and contourlet transform domain~\cite{Gho2010Three},
or patch groups regularized by group sparsity~\cite{Usman2011k,Peng2015Incorporating}.
Although these models are easy and fast to optimize, the usage of a certain sparse transform may introduce staircase artifacts or
blocky artifacts in reconstructed images~\cite{Gho2010Three}.
Combined regularization  in some of these transform domains\cite{yang2010fast}
or adaptive transform domain~\cite{Wang2016Two} can further improve reconstruction performance.
Dictionary learning method relying on a dictionary trained from reference images
has also been introduced in CS-MRI model~\cite{ravishankar2011mr,Babacan2011Reference,zhan2015fast}.
The non-local method uses groups of similar local patches for joint patch-level reconstruction to better preserve image details~\cite{Liang2011Sensitivity,qu2014magnetic,eksioglu2016decoupled}.
Methods in~\cite{JCTV2017,Huang2012Compressed} take the advantages of both the local transforms such as wavelet, TV and nonlocal similarity of the image to  improve the reconstruction quality.

In performance, the basic CS-MRI methods with traditional transformation run fast but produce less accurate reconstruction results.
Dictionary learning-based and the non-local methods generally output higher quality MR images, but suffer from slow reconstruction speed.
\subsection{Deep Learning CS-MRI Methods}
Deep learning methods are capable of extracting features from data
that are useful for recognition and restoration using end-to-end learning.
They has been applied to medical imaging to replace the traditional model-based method.
The first deep learning method for CS-MRI is in~\cite{Wang2016Accelerating}.
They trained a deep convolution network for learning the mapping from
down-sampled reconstruction images to fully sampled reconstruction images.
Method in~\cite{Schlemper2017A} applies a deep cascade of convolutional neural networks to accelerate MR imaging.
Deep residual learning for MR imaging has also been proposed.
The work in~\cite{unet} proposes a single-scale residual learning network and a multi-scale residual learning network (i.e., U-net) for image reconstruction.
They show that the U-net which is effective for image segmentation is also valid in estimation of the aliasing
artifacts.
For deep learning methods, although the reconstruction speed is fast, it is difficult to define the architecture of the network, and training a general deep network requires a large number of training samples.

\subsection{Discriminative Learning Methods}
The discriminative parameter learning approach has been applied to  sparse coding~\cite{gregor2010learning},
Markov Random Filed (MRF)~\cite{schmidt2014shrinkage,sun2015color,hershey2014deep} and non-negative matrix factorization~\cite{hershey2014deep}.
This category of approach links the conventional model-based approaches to the deep learning approaches.
The work in~\cite{gregor2010learning} implements a truncated form of coordinate
descent algorithm~\cite{Li2009Coordinate} and ISTA algorithm~\cite{Daubechies2003An} to  approximate the estimates of the sparse coding method.  Chen et al.  ~\cite{Chen2015On} and Hammernik et al.~\cite{variational2016} learn  conventional nonlinear reaction diffusion models with parameterized linear filters and parameterized influence functions to image denoising and reconstruction.
Xin et al.~\cite{Xin2016Maximal1} unfold the hard thresholding method  for a  $l_{0}$ penalized least square model to a deep neural network. They also explore the network performance of non-shared layers and shared layers.
The approaches in~\cite{schmidt2014shrinkage,sun2015color} learn the optimal shrinkage operators in Markov Random Filed model
for effective image restoration.

In this paper, we design the ADMM-Net inspired by the ADMM algorithm optimizing a general compressive sensing  model for MRI. Compared with the traditional CS-MRI methods, our approach learns optimal parameters, i.e. the parameters in the CS-MRI model and the ADMM algorithm, embedded in a deep architecture. Compared with the widely used deep neural networks, our proposed deep architectures are non-conventional in both structures and operations in network layers.
Extensive experiments show that it significantly outperforms the traditional CS-MRI models and widely utilized CNN structures in performance for MR image reconstruction. The ADMM-Net can also be potentially applied to other related problems solving an energy minimization problem using the ADMM algorithm.

\section{ADMM-Net for CS-MRI}
In this section, we first introduce a general CS-MRI  model  and corresponding ADMM
iterative procedures.
Then, we  define a  data flow graph  derived from the ADMM iterations.
Finally, we  generalize this data flow graph to construct our deep ADMM-Nets.

\subsection{General CS-MRI Model and ADMM Algorithm}
As a starting point, we consider reconstructing an MR image by a general compressive sensing MRI (CS-MRI) model.

\begin{figure*}
\centering
\includegraphics[width=0.7\linewidth]{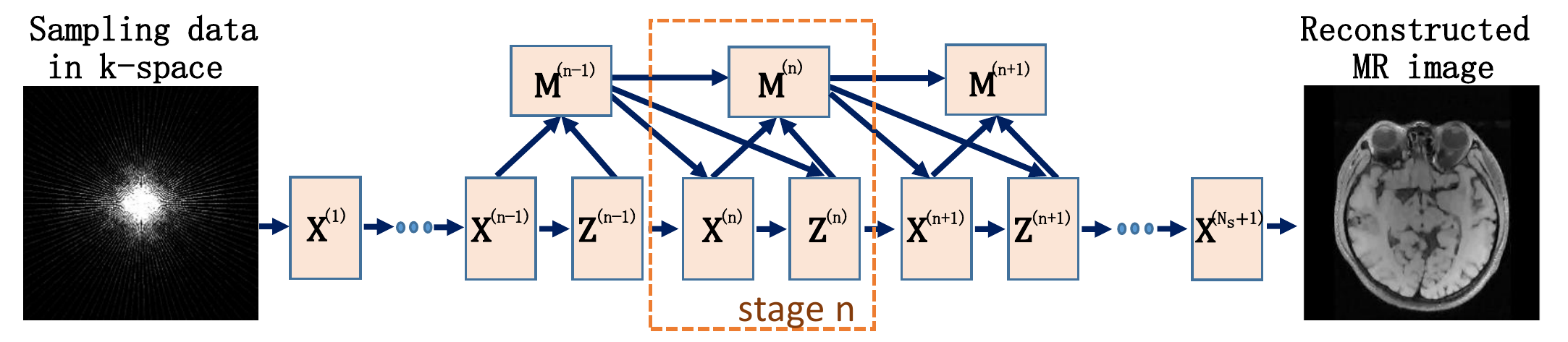}

\caption{The data flow graph for the ADMM optimization of a general CS-MRI model.
 This graph consists of three types of nodes:
reconstruction ($\bf X$), denoising ($\bf Z$), and multiplier update ($\bf M$).
An under-sampled data in $k$-space is successively processed over the graph, and finally generates an MR image.
Our deep ADMM-Nets are defined over this data flow graph.
}
\label{fig:dataflow}
\vspace{-0.2cm}
\end{figure*}

{\bf General CS-MRI Model}: Assume $x\in\mathbb{C}^{N}$ is an MRI image to be reconstructed, $y\in\mathbb{C}^{N'}$ ($N' < N$) is the under-sampled $k$-space data. According to the CS theory, the reconstructed image can be estimated by solving the following optimization problem:
\begin{equation}
\label{eq:CS}
\hat{x}=\mathop {\arg \min }\limits_x \left\{{\frac{1}{2}}\|Ax-y\|^{2}_{2}+\sum^{L}_{l=1}\lambda_{l}g(D_{l}x)\right\},
\end{equation}
where $ A=PF \in\mathbb{R}^{N'\times{N}}$ is a measurement matrix,
$P\in\mathbb{R}^{N' \times N}$ is an under-sampling matrix, $F$ is a Fourier transform.
$D_{l}$ denotes a transform matrix for a filtering operation, e.g., Discrete Wavelet Transform (DWT), Discrete Cosine Transform (DCT), etc., $g(\cdot)$ is a regularization function derived from the data prior, e.g., $l_{q}$-regularizer ($ q \in [0,1]$) for a sparse prior. $\lambda_{l}$ is  regularization parameter.

The above optimization problem can be solved efficiently by ADMM algorithm.
The following solvers are two forms of ADMM algorithm depending on the different auxiliary variables.

{\bf ADMM solver \uppercase\expandafter{\romannumeral1}}: By introducing independent auxiliary variables $z=\{z_{1}, z_{2}, \cdots, z_{L}\}$ in the transform domains,
Eqn.~(\ref{eq:CS}) is equivalent to:
\begin{equation}
\begin{split}
\mathop {\min }\limits_{x} {\frac{1}{2}}\|Ax-y\|^{2}_{2}+\sum^{L}_{l=1}\lambda_{l}g(z_{l})\;\;\;\;\;\;  \\
\;s.t.\;\; z_{l}=D_{l}x, \;\; \forall \; l \in \{1, 2, \cdots, L\}.
\end{split}
\end{equation}
Its augmented Lagrangian function is:
\begin{equation}
\begin{split}
\mathfrak{L}_{{\rho}}(x, {z}, {\alpha})=&\frac{1}{2}\left\| {Ax - y} \right\|_2^2 +
\sum_{l = 1}^L[\lambda_{l} g(z_{l})+ \\
&\langle \alpha _{l},D_{l}x - z_{l} \rangle
+\frac{\rho_{l}}{2}\|D_{l}x - z_{l}\|^{2}_{2}],
\end{split}
\end{equation}
where $\alpha = \{\alpha_{l}\}$ are Lagrangian multipliers representing  dual variables and $\rho=\{\rho_{l}\}$ are
penalty parameters.
For simplicity, we use scaled definition of  $\beta_{l}=\frac{\alpha_{l}}{\rho_{l}}$ ($l \in \{1, 2, \cdots, L\}$),
in this case, ADMM alternatively optimizes $\{x, z, \beta\}$ by solving the following three subproblems:
\begin{equation}
\begin{cases}
\begin{aligned}
   &\mathop {\arg \min }\limits_{x}{\frac{1}{2}}\|Ax-y\|^{2}_{2}+
                  \sum_{l = 1}^L \frac{\rho_{l}}{2}\|D_{l}x + \beta_{l}- z_{l}\|^{2}_{2}], \\
   &\mathop {\arg \min }\limits_{z}\sum_{l = 1}^L[\lambda_{l} g(z_{l})+\frac{\rho_{l}}{2}\|D_{l}x+
              \beta_{l}-
              z_{l}\|^{2}_{2}], \\
  &\mathop {\arg \max }\limits_{\beta}\sum_{l = 1}^L\langle\beta _{l},D_{l}x-z_{l}\rangle.
\end{aligned}
\end{cases}
\label{eq:admm1}
\end{equation}
Substitute $A=PF$ into Eqn.~(\ref{eq:admm1}), then the three subproblems have the following solutions:
\begin{equation}
\label{eq:iters1}
\begin{cases}
\begin{aligned}
&{\bf X^{(n)}}:x^{(n)}=F^{T}[P^{T}P+\sum^{L}_{l=1}\rho_{l}FD_{l}^{T}D_{l}F^{T}]^{-1}\\
&\;\;\;\;\;\;\;\;\;\;\;\;\;\;\;\;[P^{T}y+\sum^{L}_{l=1}\rho_{l}FD_{l}^{T}
(z_{l}^{(n-1)}-\beta_{l}^{(n-1)})], \\
&{\bf Z^{(n)}}\;: z_{l}^{(n)}\;=\mathcal{S}(D_{l}x^{(n)}+\beta_{l}^{(n-1)}; {\lambda_{l}}/\rho_{l}), \\
&{\bf M^{(n)}}:\beta_{l}^{(n)}=\beta_{l}^{(n-1)}+\eta_{l}(D_{l}x^{(n)}-z_{l}^{(n)}),
\end{aligned}
\end{cases}
\end{equation}
where $n \in \{1,2,\cdots, N_{s}\}$ denotes $n$-th iteration and the superscript $T$ refers to the transpose and conjugate transpose
operator for real  and complex data respectively.
Operation $x^{(n)}$ can be efficiently computed by fast Fourier transform (FFT)  because
$P^{T}P+\sum^{L}_{l=1}\rho_{l}FD_{l}^{T}D_{l}F^{T}$ is  diagonal matrix.
$\mathcal{S}(\cdot)$ is a nonlinear shrinkage function with the parameters ${\lambda_{l}}/\rho_{l}, l \in \{1, 2, \cdots, L\}$ .
It is  usually a soft or hard thresholding function corresponding to
the sparse regularization of $l_{1}$ and $l_{0}$ regularizer respectively~\cite{bach2012optimization}.
The parameter $\eta_{l}$ is an update rate for updating the  multiplier.

{\bf  ADMM solver \uppercase\expandafter{\romannumeral2}}: 
By introducing the  auxiliary variable $z$ in the spatial domain (i.e., image domain),
 Eqn.~(\ref{eq:CS}) is equivalent to:
\begin{equation}
\begin{split}
&\mathop {\min }\limits_{x,{z}} {\frac{1}{2}}\|Ax-y\|^{2}_{2}+\sum^{L}_{l=1}\lambda_{l}g(D_{l}z)\;\;\;\;\;\;  \\
&\;s.t.\;\; z=x.
\end{split}
\end{equation}
In this way,  its augmented Lagrangian function is :
\begin{equation}
\begin{split}
\mathfrak{L}_{{\rho}}(x,{z},{\alpha})=&\frac{1}{2}\left\| {Ax - y} \right\|_2^2 +
\sum_{l = 1}^L\lambda_{l}  g(D_{l}z)+\\
&\langle \alpha ,x-z \rangle
+
\frac{\rho}{2}\|x-z\|^{2}_{2}.
\end{split}
\end{equation}
Using the scaled Lagrangian multiplier $\beta = \frac{\alpha}{\rho}$, we can express the subproblems as:
\begin{equation}
\begin{cases}
\begin{aligned}
&\mathop {\arg \min }\limits_{x}{\frac{1}{2}}\|Ax-y\|^{2}_{2}-\frac{\rho}{2}\|x+\beta-z\|^{2}_{2}
                  , \\
&\mathop {\arg \min }\limits_{z} \sum_{l = 1}^L\lambda_{l}g(D_{l}z)+
\frac{\rho}{2}\|x+\beta-z\|^{2}_{2}
 ,\\
&\mathop {\arg \max }\limits_{\beta}\langle\beta ,x-z\rangle.
\end{aligned}
\end{cases}
\label{eq:admm2}
\end{equation}
One way to solve the second subproblem is to directly employ gradient-descent algorithm,
which yields the ADMM algorithm iterations:
\begin{equation}
\label{eq:iters2}
\begin{cases}
\begin{aligned}
&{\bf X^{(n)}}:x^{(n)}=F^{T}(P^{T}P+\rho I)^{-1}\\
&\;\;\;\;\;\;\;\;\;\;\;\;\;\;\;\;[P^{T}y+\rho F
(z^{(n-1)}-\beta^{(n-1)})], \\
&{\bf Z^{(n)}}\;: z^{(n,k)}\;=\mu_{1}z^{(n,k-1)} + \mu_{2}(x^{(n)}+\beta^{(n-1)})\\
&\;\;\;\;\;\;\;\;\;\;\;\;\;\;\;\; -\sum^{L}_{l=1}
\tilde{\lambda_{l}}D_{l}^{T}\mathcal{H} (D_{l}z^{(n,k-1)}), \\
&{\bf M^{(n)}}:\beta^{(n)}=\beta^{(n-1)}+\tilde{\eta}(x^{(n)}-z^{(n)}),
\end{aligned}
\end{cases}
\end{equation}
where $I$ is an unit matrix with size of $N \times N$.
$\mu_{1} = 1-l_{r}\rho$, $\mu_{2} = l_{r}\rho$, $\tilde{\lambda_{l}}=l_{r}\lambda_{l}$,
$l_{r}$ is the step size and $k \in \{1, 2, \cdots, N_{t}\}$ denotes the number of iterations in the gradient-descent method,
therefore $z^{(n)}=z^{(n,N_t)}$.
$\mathcal{H(\cdot)}$ refers to a nonlinear transform corresponding to the gradient of regularization function $g(\cdot)$.
The parameter $\tilde{\eta}$ is an update rate as well.

In CS-MRI, it commonly needs to run the ADMM algorithm in dozens of iterations to get a satisfactory reconstruction result. However, it is challenging to choose the appropriate transform $D_{l}$, and nonlinear
function $\mathcal{S(\cdot)}$ or $\mathcal{H(\cdot)}$ for general regularization function $g(\cdot)$.
Moreover, it is also non-trivial to tune the parameters $\rho_{l} $, $ \eta_l$, $\mu_{1}$, $\mu_{2}$, $\tilde{\lambda_{l}}$, etc.,  in ADMM solvers for reconstructing $k$-space data with different sampling rates.
To overcome these difficulties, we will design a data flow graph for the ADMM algorithm, over which
we can define  deep ADMM-Nets to discriminatively learn all the above transforms, functions, and parameters.

\subsection{Data Flow Graph for ADMM Algorithm}

To design our deep ADMM-Net, we first map both ADMM iterative procedures in Eqn.~(\ref{eq:iters1}) and Eqn.~(\ref{eq:iters2}) to one data flow graph. As shown in Fig.~\ref{fig:dataflow},
this graph comprises of {\em nodes} corresponding to different operations in ADMM algorithm,
and {\em directed edges} corresponding to the data flows between operations.
In this case, the $n$-th iteration of ADMM algorithm corresponds to the $n$-th stage of the data flow graph.
In each stage of the graph, there are  three types of nodes mapped from three types of operations in ADMM, i.e., reconstruction operation ($\bf X$) for restoring an image from Fourier transform domain,
denoising operation ($\bf Z$) regarded as denoising the image $D_{l}x+\beta_{l}, l \in \{1,2,\cdots, L\}$
in Eqn.~(\ref{eq:admm1}) or $ x + \beta$ in Eqn.~(\ref{eq:admm2}),
  and multiplier update operation ($\bf M $).
The whole data flow graph  is a multiple repetition of the above stages corresponding to successive iterations in ADMM. Given an under-sampled data in $k$-space, it flows over the graph and finally generates a reconstructed image.
In this way, we map the ADMM iterations to a data flow graph, which is useful to define and train our deep ADMM-Nets in the following. 

\subsection{Deep ADMM-Nets}

Our deep ADMM-Nets are defined over the data flow graph. They keep the graph structure but generalize and decompose the three types of operations to have learnable parameters as network layers. These  operations are now generalized as reconstruction layer,
convolution layer, non-linear transform layer, multiplier update layer, and so on. We next discuss them in details.

\subsubsection {Basic-ADMM-Net}

\begin{figure*}[!htbp]
\centering
\includegraphics[width=0.8\linewidth]{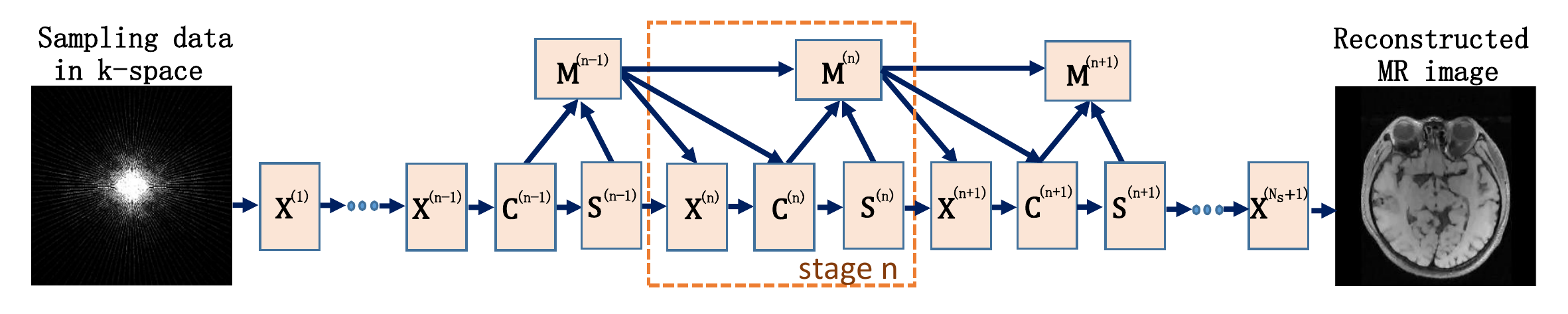}
\caption{The architecture of a deep Basic-ADMM-Net. There are four layers including reconstruction layer ($\bf X^{(n)}$), convolution layer ($\bf C^{(n)}$), nonlinear transform layer ($\bf S^{(n)}$) and multiplier update layer ($\bf M^{(n)}$) in each stage.
}
\label{fig:ADMM-Net1}
\vspace{-0.2cm}
\end{figure*}

Based on the iterative algorithm in ADMM solver I, i.e., Eqn.~(\ref{eq:iters1}), we define a deep network called Basic-ADMM-Net shown in Fig.~\ref{fig:ADMM-Net1}.
It generalizes the reconstruction operation $\bf X^{(n)}$ to  reconstruction layer,
decomposes the denoising operation $\bf Z^{(n)}$ to convolution layer derive from the  operation
$\{D_l x^{(n)}\}_{l=1}^L$ and non-linear transform layer,
finally extends the multiplier update procedure $\bf M^{(n)}$ in Eqn.~(\ref{eq:iters1}) to  multiplier update layer.

\quad {\bf {Reconstruction layer}} ($\bf X^{(n)}$): This layer reconstructs an MRI image following the reconstruction operation $\bf X^{(n)}$ in Eqn.~(\ref{eq:iters1}). Given the inputs $z_{l}^{(n-1)}$ and $\beta_{l}^{(n-1)}$ , the output of this layer is defined as:
\begin{equation}
\begin{aligned}
x^{(n)}=&F^{T}(P^{T}P+\sum_{l=1}^{L}\rho_{l}^{(n)}F{{H_{l}^{(n)}}}{}^{T}H_{l}^{(n)}F^{T})^{-1}\\
&[P^{T}y+\sum_{l=1}^{L}\rho_{l}^{(n)}F{H_{l}^{(n)}}{}^{T}(z_{l}^{(n-1)}-\beta_{l}^{(n-1)})],
\end{aligned}
\end{equation}
where $H_{l}^{(n)}$ is the $l$-th filter with size of $w_{f}\times w_{f}$ in stage $n$, $\rho_{l}^{(n)}$ is the $l$-th penalty parameter, $l=1, \cdots, L$, and $y$ is the input under-sampled data in $k$-space.
In the first stage (i.e., $n=1$), $z_{l}^{(0)}$ and $\beta_{l}^{(0)}$ are initialized to zeros, therefore
$ x^{(1)} = F^{T}(P^{T}P+\sum_{l=1}^{L}\rho_{l}^{(1)}FH_{l}^{(1)}{}^{T}H_{l}^{(1)}F^{T})^{-1}(P^{T}y)$.


\quad {\bf {Convolution layer}} ($\bf C^{(n)}$): It performs convolution operation to transform an image into transform domain.
To  represent  images, a  traditional choice is using  a set of pre-trained bases such as DCT, Fourier, Haar, etc. In our formulation, we represent  the image by a set of learnable filters equivalently.
Given an image $x^{(n)}$, i.e., a reconstructed image in $n$-th stage, the output is
\begin{equation}
c_{l}^{(n)}=D_{l}^{(n)} x^{(n)},
\end{equation}
where $D_{l}^{(n)}$ is a matrix corresponding to the $l$-th filter with size of  $w_{f} \times w_{f}$  in stage $n$. Different from the original ADMM, we do not constrain the filters $H^{(n)}_l$ and $D^{(n)}_l$ to be the same to increase the network capacity.
\begin{figure}[h]
\centering
\includegraphics[height=1.8cm]{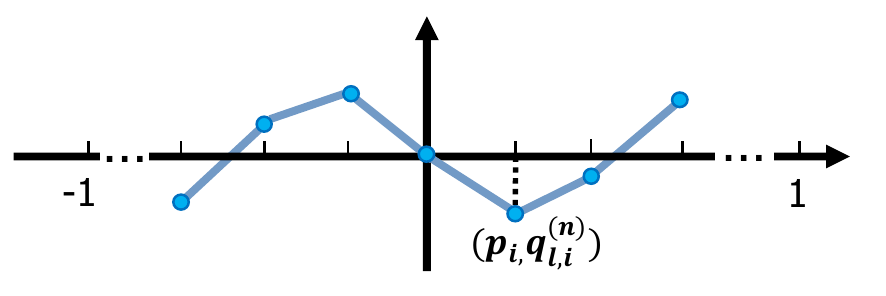}
\caption{Illustration of a piecewise linear function determined by a set of control points$(p_i, q^{(n)}_{l, i})$.
}
\label{fig:nonliear}
\end{figure}

\quad {\bf {Nonlinear transform layer}} ($\bf S^{(n)}$): This layer performs nonlinear transform inspired by the shrinkage function $\mathcal{S}(\cdot)$ defined in $\bf Z^{(n)}$ from Eqn.~(\ref{eq:iters1}). Instead of setting it to be a shrinkage function determined by the regularization term $g(\cdot)$ in Eqn.~(\ref{eq:CS}), we aim to learn more general function using piecewise linear function. Given $c_{l}^{(n)}$ and $\beta_{l}^{(n-1)}$, the output of this layer is defined as:
\begin{equation}
z_{l}^{(n)}=S_{PLF}(c_{l}^{(n)}+\beta_{l}^{(n-1)};  \{p_i, q^{(n)}_{l, i}\}^{N_c}_{i =1}),
\end{equation}
where $S_{PLF}(\cdot)$ is a piecewise linear function determined by a set of control points $\{p_i, q^{(n)}_{l,i}\}_{i=1}^{N_c}$.  i.e.,
\begin{equation}
\nonumber
S_{PLF}(a)=
\left\{
  \begin{array}{ll}
    a+q_{l,1}^{(n)}-p_{1}, & \hbox{$a < p_{1}$,} \\
    a+q_{l,N_{c}}^{(n)}-p_{N_{c}}, & \hbox{$a > p_{N_{c}}$,} \\
    q_{l,r}^{(n)}+\frac{(a-p_{r})(q_{l,r+1}^{(n)}-q_{l,r}^{(n)})}{p_{r+1}-p_{r}},&\hbox{$p_{1}\leq a \leq p_{N_{c}}$,}
  \end{array}
\right.
\end{equation}
where $r=\lfloor\frac{a-p_{1}}{p_{2}-p_{1}}\rfloor$,
$\{p_i\}_{i=1}^{N_c}$ are predefined positions uniformly located within
[-1,1], and $\{q^{(n)}_{l,i}\}_{i=1}^{N_c}$ are the values at these positions for $l$-th filter in $n$-th stage. Figure~\ref{fig:nonliear} gives an illustrative example.
Since a piecewise linear function
can approximate any function, we can learn flexible nonlinear transform function from data beyond the off-the-shelf hard or soft thresholding functions.


\quad {\bf {Multiplier update layer}} ($\bf M^{(n)}$): Given three inputs $\beta_{l}^{(n-1)}$, $c_{l}^{(n)}$ and $z_{l}^{(n)}$, the output of this layer is defined as:
\begin{equation}
\beta_{l}^{(n)}=\beta_{l}^{(n-1)}+\eta_{l}^{(n)}(c_{l}^{(n)}-z_{l}^{(n)}),
\end{equation}
where $\eta_l^{(n)}$ is a learnable parameter.

\quad {\bf {Network Parameters}}: In this deep architecture, we aim to learn the following parameters:
$H_l^{(n)}$ and $\rho_{l}^{(n)}$ in reconstruction layer, filter
$D^{(n)}_l$  in convolution layer,  the values $\{ q^{(n)}_{l,i} \}^{N_c}_{i=1}$  in nonlinear transform layer, and $\eta_l^{(n)}$ in multiplier update layer, where $l \in \{1, 2, \cdots, L\}$ and $n \in \{1, 2, \cdots, N_s\}$ are the indexes for the filters and stages respectively. All of these parameters are taken as the network parameters to be learned.

\begin{figure*}[!htbp]
\vspace{-0.2cm}
\centering
\includegraphics[width=0.6\linewidth]{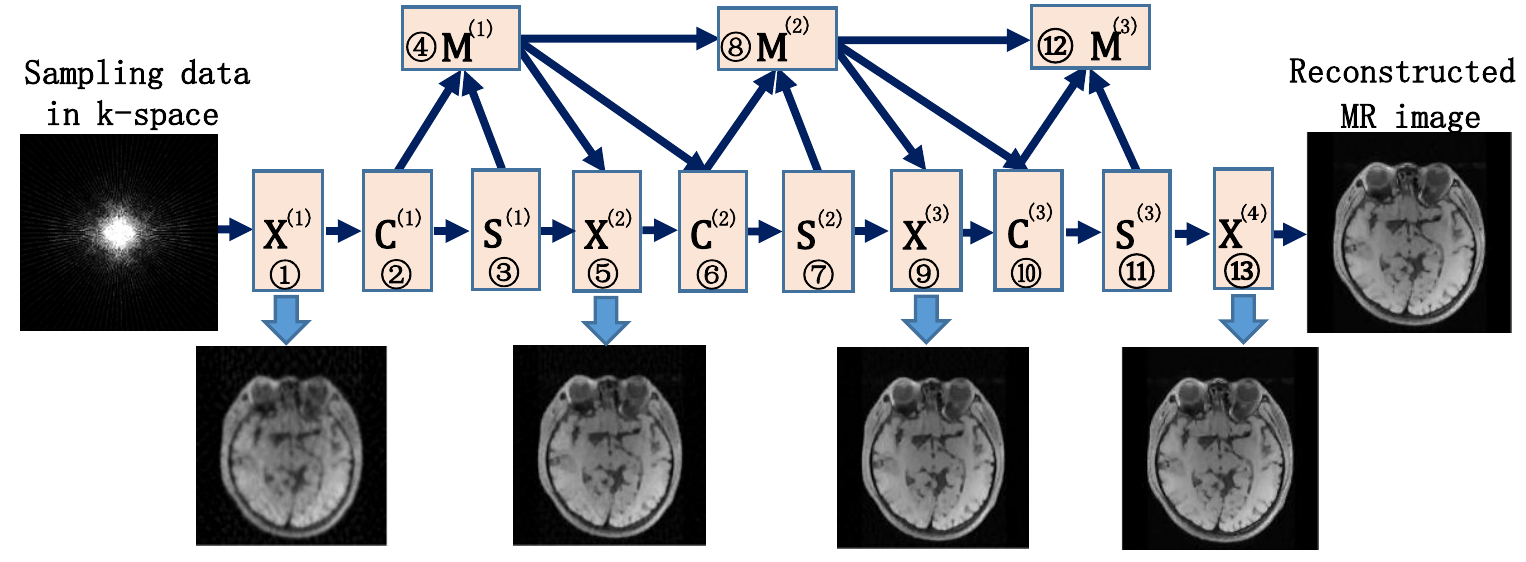}

\caption{An example of a deep Basic-ADMM-Net with three stages. The sampled data in $k$-space is successively processed by operations in a order from 1 to 12, followed by a reconstruction layer $X^{(4)}$ to output the final reconstructed image. The reconstructed image in each stage is shown under each reconstruction layer.
}
\label{fig:ADMM-Net1ex1}
\vspace{-0.2cm}
\end{figure*}

Figure~\ref{fig:ADMM-Net1ex1} shows an example of a deep Basic-ADMM-Net with three stages. The under-sampled data in $k$-space flows over three stages in a order
from circled number 1 to number 12, followed by a final reconstruction layer with circled number 13 and generates a reconstructed MR image. Immediate reconstruction result at each stage is shown under each reconstruction layer.
\subsubsection {Generic-ADMM-Net}
\begin{figure*}[!htbp]
\centering
\includegraphics[width=0.65\linewidth]{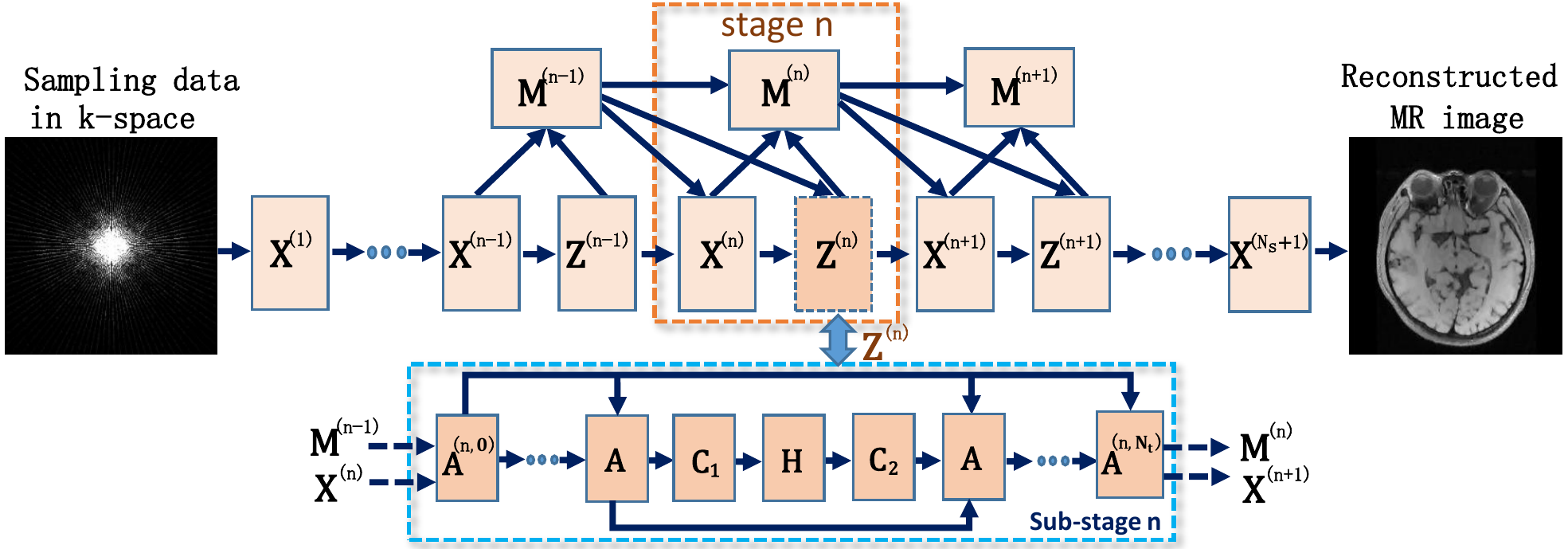}

\caption{The architecture of a deep Generic-ADMM-Net. There are three layers including reconstruction layer ($\bf X^{(n)}$), denoising layer ($\bf Z^{(n)}$) and multiplier update layer ($\bf M^{(n)}$) in each stage.
The denoising layer $\bf Z^{(n)}$ is decomposed to $N_{t}$ iterations  consisting of  addition layer ($\bf A$), convolution layers ($\bf C_{1},C_{2}$) and non-linear transform layer ($\bf H$) in each sub-stage.
}
\label{fig:Z}
\vspace{-0.2cm}
\end{figure*}

From the iterative operations in ADMM solver II, i.e.,  Eqn.~(\ref{eq:iters2}), we extent the data flow graph to a deep
Generic-ADMM-Net as shown in Fig.~\ref{fig:Z}.
Each stage of the network  corresponds to one iteration of the ADMM algorithm in Eqn.~(\ref{eq:iters2}), and each sub-stage of the network  corresponds to iterations of the  gradient-descent algorithm implementing  $Z^{(n)}$   in  Eqn.~(\ref{eq:iters2}). Compared with the Basic-ADMM-Net,
the Generic-ADMM-Net has a different reconstruction layer  $\bf X^{(n)}$ and a generalized denoising layer $\bf Z^{(n)}$ in each stage.

\quad {\bf {Reconstruction layer}} ($\bf X^{(n)}$):  Given $z^{(n-1)}$ and $\beta^{(n-1)}$, the output of this layer is defined as:
\begin{equation}
\begin{aligned}
x^{(n)}=&F^{T}(P^{T}P+\rho^{(n)}I)^{-1}\\
&[P^{T}y+\rho^{(n)}F(z^{(n-1)}-\beta^{(n-1)})],
\end{aligned}
\end{equation}
where $\rho^{(n)}$ is a learnable penalty parameter and $y$ is the input under-sampled data in $k$-space.
In the first stage (i.e., $n=1$), $z_{l}^{(0)}$ and $\beta_{l}^{(0)}$ are also initialized to zeros, therefore
$ x^{(1)} = F^{T}(P^{T}P+\rho^{(1)}I)^{-1}(P^{T}y).$

We next introduce the denoising layer $\bf Z^{(n)}$.
We decompose  $\bf Z^{(n)}$ in Eqn.~(\ref{eq:iters2}) to  several sub-layers, such as  addition layer,  convolution layer and non-linear transform layer. The iterative stacking of these sub-layers  just implement the iterative gradient descent in  $\bf Z^{(n)}$ of Eqn.~(\ref{eq:iters2}). For the $k$-th iteration ($k \in \{1,2,\cdots, N_{t}\}$) in the $n$-th sub-stage  in Fig.~\ref{fig:Z}, these sub-layers are defined as follows.

\quad {\bf {Addition layer}} ($\bf A^{(n,k)}$): It performs simple weighted summation operation. Given  $x^{(n)}$, $\beta^{(n-1)}$,
 $z^{(n,k-1)}$ and $c^{(n,k)}_{2}$, the output of this layer is
 \begin{equation}
  z^{(n,k)}= \mu_{1}^{(n,k)}z^{(n,k-1)}+\mu_{2}^{(n,k)}(x^{(n)}+\beta^{(n-1)})- c_{2}^{(n,k)},
 \end{equation}
where $\mu_{1}^{(n,k)}$ and $ \mu_{2}^{(n,k)} $are weighting parameters.
$z^{(n,0)}$  is initialized to $x^{(n)}+\beta^{(n-1)}$,  therefore
$ z^{(n,1)}=x^{(n)}+\beta^{(n-1)}-c_{2}^{(n,1)}$ due to $\mu_{1}+\mu_{2}=1$ in Eqn.~(\ref{eq:iters2}). The output of the $n$-th sub-stage is $z^{(n)}=z^{(n,N_t)}$.

\quad {\bf {Convolution layer}} ($\bf C_{1}^{(n,k)},C_{2}^{(n,k)}$):
Given the input $z^{(n,k-1)}$ and $h^{(n,k)}$,
we define two convolution layers as:
\begin{equation}
c_{1}^{(n,k)}=w_{1}^{(n,k)}\ast z^{(n,k-1)}+b_{1}^{(n,k)},
\end{equation}
\begin{equation}
c_{2}^{(n,k)}=w_{2}^{(n,k)}\ast h^{(n,k)}+b_{2}^{(n,k)},
\end{equation}
where $w_{1}^{(n,k)}$ corresponds to $L$ filters with size of $w_f \times w_f $, $b_{1}^{(n,k)}$ is
an $L$-dimensional biases vector, $w_{2}^{(n,k)}$ corresponds to a filter with  size of $f \times f \times L $, and $b_{2}^{(n,k)}$ is
a $1$-dimensional biases vector.
We  integrate feature extraction and feature fusion in the second convolution operation,
the output $c_{1}^{(n,k)}$ is composed of $L$ feature maps and the output $c_{2}^{(n,k)}$ is a feature map which has  the same size as the  image  to be reconstructed, therefore we generalize
the operation $\sum^{L}_{l=1}
\tilde{\lambda_{l}}D_{l}^{T}\mathcal{H} (D_{l}z^{(n,k-1)})$ of Eqn.~(\ref{eq:iters2}) to the operation $ w_{2}^{(n,k)}\ast \mathcal{H}(w_{1}^{(n,k)}\ast z^{(n,k-1)}+b_{1}^{(n,k)})+b_{2}^{(n,k)} = c_{2}^{(n,k)}$.

\quad {\bf {Nonlinear transform layer}} ($\bf H^{(n,k)}$): We still learn more general function using piecewise linear function for $\mathcal{H}(\cdot)$ defined in $\bf Z^{(n)}$ in Eqn.~(\ref{eq:iters2}). Given the input $c_{1}^{(n,k)}$ , the output of this layer is defined as:
\begin{equation}
h^{(n,k)}=S_{PLF}(c_{1}^{(n,k)};  \{p_i, q^{(n,k)}_{i}\}^{N_c}_{i =1}),
\end{equation}
 where $\{p_i\}_{i=1}^{N_c}$ are predefined positions uniformly located within
[-1,1], and $\{q^{(n,k)}_{i}\}_{i=1}^{N_c}$ are the values at these positions  for $k$-th iteration in the
sub-stage $n$.

Generic-ADMM-Net also generalizes  the  multiplier update procedure $\bf M^{(n)}$ in Eqn.~(\ref{eq:iters2}) to  multiplier update layer.

\quad {\bf {Multiplier update layer}} ($\bf M^{(n)}$): The output of this layer in stage $n$ is defined as:
\begin{equation}
\beta^{(n)}=\beta^{(n-1)}+\tilde{\eta}^{(n)}(x^{(n)}-z^{(n)}),
\end{equation}
where $\tilde{\eta}^{(n)}$ is a learnable parameter in $n$-th stage.

\begin{figure*}[!htbp]
\centering
\includegraphics[width=0.9\linewidth]{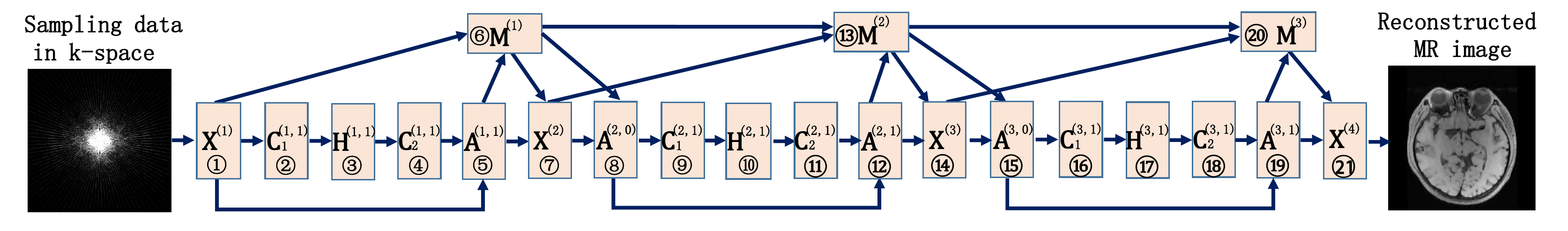}

\caption{An example of deep Generic-ADMM-Net with three stages and one iteration in each sub-stage ($n=3$,$N_{t}=1$). The sampled data in $k$-space is successively processed by operations in an order from 1 to 20, followed by a reconstruction layer $X^{(4)}$ to output the final reconstructed image.
}
\label{fig:ADMM-Net2ex1}
\vspace{-0.2cm}
\end{figure*}

\quad {\bf {Network Parameters}}: In this deep architecture, we aim to learn the following parameters:
$\rho^{(n)}$ in  reconstruction layer, $\mu_{1}^{(n,k)}$ and $\mu_{2}^{(n,k)}$ in   addition layer,
filters $w^{(n,k)}_{1}$ , $w^{(n,k)}_{2}$, biases $b^{(n,k)}_{1}$ , $b^{(n,k)}_{2}$ in convolution layers,
$\{ q^{(n,k)}_{i} \}^{N_c}_{i=1}$  in nonlinear transform layer and $\tilde{\eta}^{(n)}$ in  multiplier update layer,
where $n \in \{1,2,\cdots, N_s\}$ and $k \in \{1,2,\cdots,N_{t}\}$   index the stages and the iterations in a sub-stage respectively. Figure~\ref{fig:ADMM-Net2ex1} shows an example of a Generic-ADMM-Net  with three stages and one iteration in each sub-stage.

\subsubsection {Complex-ADMM-Net}

Generally, MR images are complex-valued images, and their phase information is also important for medical applications.
Unlike many other popular deep networks  designed for applications processing real-valued data,
we generalize our Generic-ADMM-Net to  reconstruct  complex-valued MR images,  dubbed Complex-ADMM-Net.
Compared with Generic-ADMM-Net, the Complex-ADMM-Net has a generalized operation in the nonlinear transform layer.

For the convolution layers, we still use real-valued filters to convolve complex-valued data. It means that the filters in convolution layers are all  real-valued, and the convolution is performed between the real-valued filters  and complex-valued inputs. This is equivalently to performing convolution over real and imagery parts of layer input separately using shared real-valued filters.

For the nonlinear transform layer, we deal with real and imaginary parts separately, i.e.,
\begin{equation}
h^{(n,k)}=S_{PLF}(\mathbf{Re}(c_{1}^{(n,k)}))+jS_{PLF}(\mathbf{Im}(c_{1}^{(n,k)})),
\end{equation}
where $\mathbf{Re}(\cdot)$ is an operator for the real part and $\mathbf{Im}(\cdot)$  is an operator for the imaginary part. Both the real and imagery parts share the same  piecewise linear transform function.

\section{Network Training}

We take the  reconstructed MR image from fully sampled data in $k$-space as the ground-truth MR image $x^{gt}$, and under-sampled data $y$ in $k$-space as the input. Then a training set $\Gamma$ is constructed containing pairs of under-sampled data and ground-truth MR image.
We choose normalized mean square error (NMSE) as the loss function in network training. Given pairs of training data, the loss between the network output and the ground-truth is defined as:
\begin{equation}
E(\Theta)=\frac{1}{|\Gamma|}\sum_{(y, x^{gt}) \in \Gamma}\frac{{\|\hat{x}(y, \Theta)-x^{gt}\|_{2}}}{{\|x^{gt}\|_{2}}},
\end{equation}
where $\hat{x}(y,\Theta)$ is the network output based on network parameter $\Theta$ and under-sampled data $y$ in $k$-space. The NMSE is a widely-used evaluation criteria for image reconstruction, and ADMM-Nets  also permit usage of other kinds of evaluation metrics such as SSIM, PSNR, SNR,  etc.
We learn the  parameters
$\Theta = \{ H_{l}^{(n)}, \rho_{l}^{(n)}, D_{l}^{(n)}, \{q^{(n)}_{l,i}\}_{i=1}^{N_c}, \eta_{l}^{(n)}  \}_{n=1}^{N_{s}}\cup$
$\{ H_{l}^{(N_s+1)},\rho_{l}^{(N_s+1)}\}$\;\;$(l$$ $=$ 1,$$ \cdots, L)$
for Basic-ADMM-Net, and parameters\;
$\Theta=\{ \rho^{(n)},\mu_{1}^{(n,k)},\mu_{2}^{(n,k)}, w_{1}^{(n,k)},b_{1}^{(n,k)},w_{2}^{(n,k)},$
$b_{2}^{(n,k)}$$, \{q^{(n,k)}_{i}\}_{i=1}^{N_c}, \tilde{\eta}^{(n)}  \}_{n=1}^{N_{s}} $$\cup$$ \{ \rho^{(N_s+1)}\} $
in  Generic-ADMM-Net
by minimizing the loss w.r.t. these parameters using gradient-based  algorithm  L-BFGS~\footnote{{http://users.eecs.northwestern.edu/~nocedal/lbfgsb.html}}.
In the following, we first discuss the initialization of  parameters and then compute the gradients of the loss  $E(\Theta)$ w.r.t. parameters $\Theta$ using back-propagation (BP) over the networks.

\subsection{Initialization }
\label{sec:int}
We initialize the network parameters by model-based initialization method and random initialization method.

{\bf{Model-based initialization}}: We initialize parameters $\Theta$ in the ADMM-Nets according to the ADMM solvers of the following baseline CS-MRI model
\begin{equation}
\mathop {\arg \min }\limits_x \left\{{\frac{1}{2}}\|Ax-y\|^{2}_{2}+\lambda \sum^{L}_{l=1}||D_{l}x||_1\right\}.
\label{eqn:init}
\end{equation}
In this model, we set $D_{l}$ as a DCT basis  and impose $l_1$-norm regularizer in the DCT transform space.
Then, the function $S(\cdot)$ in ADMM algorithm (Eqn.~(\ref{eq:iters1}))  is a soft thresholding function: $S(a;{\lambda}/\rho_{l}) = sgn(a)(|a| - \lambda/\rho_{l})$ when $|a| > \lambda/\rho_{l}$, and 0 otherwise.
For each $n$-th stage of the networks,
filters $D^{(n)}_l$, $w^{(n)}_l$, $w_{1}^{(n,k)}$ and $w_{2}^{(n,k)}$
are initialized to be $D_l$ ($l=1 ,\cdots,L$) from Eqn.~(\ref{eqn:init}).
For the nonlinear transform layers,
we uniformly choose 101  positions located within [-1,1], each  value   $q^{(n)}_{l,i}$  is initialized as $S( p_i;\lambda/\rho_{l})$.
Other parameters  are initialized to be the corresponding values in  the ADMM algorithm.
In this case, the initialized net is exactly a realization of ADMM optimizing process, and the optimization of the network parameters is expected to produce improved reconstruction result.

{\bf{Random initialization}}: In the deep learning literatures, the weights in deep networks are
generally initialized by random values. In such initialization,
for $n$-th stage of the ADMM-Nets, filters are initialized
by values randomly sampled from a Gaussian distribution with zero mean and well chosen variance level~\cite{He2015Delving}.
Each value of the predefined positions (i.e., $q^{(n)}_{i}$) in a non-linear transform layer is initialized as $max( p_i, 0)$ implementing a rectified linear unit (ReLU) function. 

\begin{figure*}[!htbp]
\centering
\includegraphics[width=0.9\linewidth]{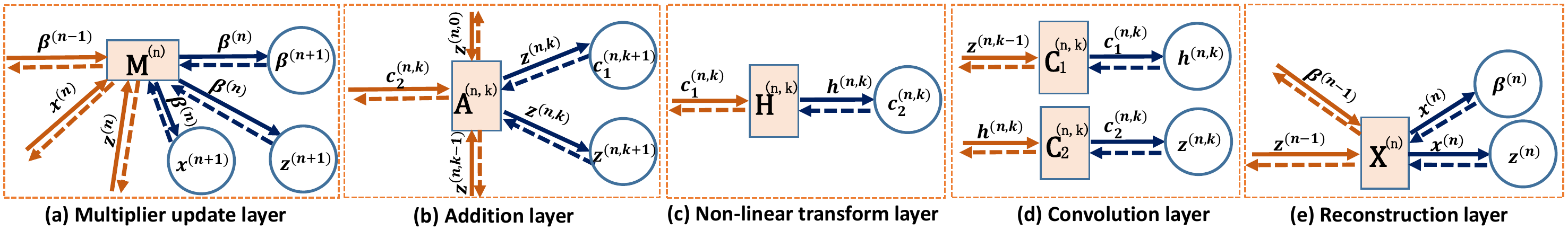}

\caption{Illustration of five types of graph nodes (i.e., layers in network) and their data flows in stage $n$. The solid arrow indicates the data flow in forward pass and dashed arrow indicates the backward pass when computing gradients in back-propagation. }
\label{fig:BP}
\vspace{-0.2cm}
\end{figure*}

\subsection{Gradient Computation by Back-propagation }

We compute the gradients of loss w.r.t. parameters using back-propagation over the deep architecture in Fig.~\ref{fig:dataflow}.
In the forward pass, we process the data of $n$-th stage in the order of $\bf X^{(n)}$,  $\bf Z^{(n)}$, $\bf M^{(n)}$.
In the backward pass, the gradients are computed in an inverse order.  Figures~\ref{fig:ADMM-Net1ex1} and ~\ref{fig:ADMM-Net2ex1} show the examples, where the gradients  can be computed backwardly from the layers with circled number 13 (or 21) to 1 successively.
For $n$-th stage in Generic-ADMM-Net, Fig.~\ref{fig:BP} shows five types of nodes (i.e., network layers) and the data flow over them. Each node has multiple inputs and (or) outputs.
We next briefly introduce the gradients computation for each layer in a typical stage $n$ 
of  Generic-ADMM-Net.
The method of gradient computation in the Basic-ADMM-Net is similar as Generic-ADMM-Net.
Please refer to supplementary material for more details.

\quad \emph{\bf Multiplier update layer  ($\bf M^{(n)}$)}: As shown in Fig.~\ref{fig:BP}(a), this layer has three inputs: $\beta^{(n-1)}, x^{(n)}$ and $z^{(n)}$. Its output  $\beta^{(n)}$ is the input to compute $\beta^{(n+1)}, z^{(n+1)}$ and $ x^{(n+1)}$. The parameter of this layer is $\tilde{\eta}^{(n)}$. The gradient of loss w.r.t. the parameter can be computed as:
\begin{equation}
\nonumber \frac{\partial E}{\partial \tilde{\eta}^{(n)}}=\frac{\partial E}{\partial \beta^{(n)T}}\frac{\partial \beta^{(n)}}{\partial \tilde{\eta}^{(n)}}
= \frac{\partial E}{\partial \beta^{(n)T}}(x^{(n)}-z^{(n)}),
\end{equation}
where $\frac{\partial E}{\partial \beta^{(n)T}}=\frac{\partial E}{\partial \beta^{(n+1)T}}\frac{\partial \beta^{(n+1)}}{\partial \beta^{(n)}}+\frac{\partial E}{\partial z^{(n+1)T}}\frac{\partial z^{(n+1)}}{\partial \beta^{(n)}}+\frac{\partial E}{\partial x^{(n+1)T}}$
$\frac{\partial x^{(n+1)}}{\partial \beta^{(n)}}$,
is the summation of gradients along  three dashed blue arrows in Fig.~\ref{fig:BP}(a).
We also compute gradients of the loss w.r.t. its inputs in this layer for gradient propagation:
\begin{equation}
\nonumber \frac{\partial E}{\partial \beta^{(n-1)}}=\frac{\partial E}{\partial \beta^{(n)T}}\frac{\partial \beta^{(n)}}{\partial \beta^{(n-1)}}
= \frac{\partial E}{\partial \beta^{(n)T}},
\end{equation}
\begin{equation}
\nonumber \frac{\partial E}{\partial x^{(n)}} = \tilde{\eta}^{(n)}\frac{\partial E}{\partial \beta^{(n)T}},\;\;
 \frac{\partial E}{\partial z^{(n)}} =-\frac{\partial E}{\partial x^{(n)}} .
\end{equation}

\quad \emph{\bf Addition layer ($\bf A^{(n,k)}$)}:
The parameters of this layer are $\mu_{1}^{(n,k)}$ and
$\mu_{2}^{(n,k)}$. As shown in Fig.~\ref{fig:BP}(b), the output $z^{(n,k)}$ is the input for computing $c_{1}^{(n,k+1)}$ and $z^{(n,k+1)}$ in the next layers. The gradients of  loss w.r.t. parameters can be computed as
\begin{equation}
\nonumber \frac{\partial E}{\partial \mu_{1}^{(n,k)}}=\frac{\partial E}{\partial z^{(n,k)T}}z^{(n,k-1)},\;\;
\frac{\partial E}{\partial \mu_{2}^{(n,k)}}=\frac{\partial E}{\partial z^{(n,k)T}},
\end{equation}
where $\frac{\partial E}{\partial z^{(n,k)T}}=\frac{\partial E}{\partial c_{1}^{(n,k+1)T}}\frac{\partial c_{1}^{(n,k+1)}}{\partial z^{(n,k)}}+
\frac{\partial E}{\partial z^{(n,k+1)T}}\frac{\partial z^{(n,k+1)}}{\partial z^{(n,k)}}.
$
The gradients of the loss w.r.t. the inputs in this layer are:
\begin{equation}
\nonumber \frac{\partial E}{\partial z^{(n,k-1)}} = \mu_{1}^{(n,k)}\frac{\partial E}{\partial z^{(n,k)T}},\;\;
\nonumber \frac{\partial E}{\partial c_{2}^{(n,k)}} = -\frac{\partial E}{\partial z^{(n,k)T}},
\end{equation}
\begin{equation}
\nonumber \frac{\partial E}{\partial x^{(n)}} =\frac{\partial E}{\partial \beta^{(n-1)}}= \mu_{2}^{(n,k)}\frac{\partial E}{\partial z^{(n,k)T}}.
\end{equation}

\quad \emph{\bf Nonlinear transform layer ($\bf H^{(n,k)}$)}: As shown in Fig.~\ref{fig:BP}(c), this layer has
one input: $c_{1}^{(n,k)}$, and its output $h^{(n,k)}$ is the input for computing $c_2^{(n,k)}$.
The parameters of this layer are $\{q^{(n,k)}_{i}\}_{i=1}^{N_{c}}$.
The gradient of  loss w.r.t. parameters can be computed as
\begin{equation}
\nonumber
\frac{\partial E}{\partial q_{ i}^{(n,k)}}=\frac{\partial E}{\partial h^{(n,k)T}}\frac{\partial h^{(n,k)}}{\partial q_{ i}^{(n,k)}},
\end{equation}
where $\frac{\partial E}{\partial h^{(n,k)T}}=\frac{\partial E}{\partial c_{2}^{(n,k)T}}\frac{\partial c_{2}^{(n,k)}}{\partial h^{(n,k)}}$.
The calculation of   $\frac{\partial h^{(n,k)}}{\partial q_{ i}^{(n,k)}}$  and
$\frac{\partial h^{(n,k)}}{\partial c_{1}^{(n,k)}}$  can be found in the supplementary material.

\quad \emph{\bf Convolution layer ($\bf C_{1}^{(n,k)},\bf C_{2}^{(n,k)}$)}: As shown in Fig.~\ref{fig:BP}(d),
a convolution layer has one input and one output.
The parameters of $\bf C_{1}^{(n,k)}$ layer and $\bf C_{2}^{(n,k)}$ layer are $w_{1}^{(n,k)}, b_{1}^{(n,k)}, w_{2}^{(n,k)}$ and $b_{2}^{(n,k)}$.
The gradient of  loss w.r.t. parameters can be computed as
\begin{equation}
\nonumber
\frac{\partial E}{\partial w_{1}^{(n,k)}}=\frac{\partial E}{\partial c_{1}^{(n,k)T}}\frac{\partial c_{1}^{(n,k)}}{\partial w_{1}^{(n,k)}},\;\;
\frac{\partial E}{\partial b_{1}^{(n,k)}}=\frac{\partial E}{\partial c_{1}^{(n,k)T}},
\end{equation}
\begin{equation}
\nonumber
\frac{\partial E}{\partial w_{2}^{(n,k)}}=\frac{\partial E}{\partial c_{2}^{(n,k)T}}\frac{\partial c_{2}^{(n,k)}}{\partial w_{2}^{(n,k)}},\;\;
\frac{\partial E}{\partial b_{2}^{(n,k)}}=\frac{\partial E}{\partial c_{2}^{(n,k)T}}.
\end{equation}
Where $\frac{\partial E}{\partial c_{1}^{(n,k)T}}=\frac{\partial E}{\partial h^{(n,k)T}}\frac{\partial h^{(n,k)}}{\partial c_{1}^{(n,k)}}$,
$\frac{\partial E}{\partial c_{2}^{(n,k)T}}=\frac{\partial E}{\partial z^{(n,k)T}}\frac{\partial z^{(n,k)}}{\partial c_{2}^{(n,k)}}$.
We also compute gradients of the loss w.r.t. its inputs in this layer for back propagation:
\begin{equation}
\nonumber
\frac{\partial E}{\partial z^{(n,k-1)}}=\frac{\partial E}{\partial c_{1}^{(n,k)T}}\frac{\partial c_{1}^{(n,k)}}{\partial z^{(n,k-1)}},
\frac{\partial E}{\partial h^{(n,k)}}=\frac{\partial E}{\partial c_{2}^{(n,k)T}}\frac{\partial c_{2}^{(n,k)}}{\partial h^{(n,k)}}.
\end{equation}


\quad \emph{\bf Reconstruction layer ($\bf X^{(n)}$)}: The parameter in this layer is $\rho^{(n)}$ for  $n$-th stage. The gradient of loss w.r.t.  the parameter is computed as
\begin{equation}
\begin{split}
\nonumber \frac{\partial E}{\partial \rho^{(n)}}=
&\frac{\partial E}{\partial x^{(n)T}}
F^{T}Q\{F(z^{(n-1)}-\beta ^{(n-1)})\\
&-Q[P^{T}y+\rho^{(n)}F(z^{(n-1)}-\beta ^{(n-1)})]
\},
\end{split}
\end{equation}
where $Q=(P^{T}P+\rho^{(n)}I)^{-1}$, and $\frac{\partial E}{\partial x^{(n)T}}$ is
\begin{equation}
\nonumber
 \frac{\partial E}{\partial x^{(n)T}}=
\left\{
  \begin{array}{ll}
    \frac{\partial E}{\partial \beta^{(n)T}}\frac{\partial \beta^{(n)}}{\partial x^{(n)}}+\frac{\partial E}{\partial z^{(n)T}}\frac{\partial z^{(n)}}{\partial x^{(n)}},&\hbox{$n\leq N_{s}$,} \\
    \frac{1}{|\Gamma|} \frac{(x^{(n)}-x^{gt})}{{\|x^{gt}\|_{2}}{\|x^{(n)}-x^{gt}\|_{2}}},&\hbox{$n=N_{s}+1$.}
  \end{array}
\right.
\end{equation}
The gradient of loss w.r.t. input in this layer is computed as
\begin{equation}
\nonumber \frac{\partial E}{\partial z^{(n-1)}} =
[\rho^{(n)}F^{T}QF\frac{\partial E}{\partial x^{(n)}}]^{T}
= -\frac{\partial E}{\partial \beta^{(n-1)}}.
\end{equation}

\quad \emph{\bf Complex-ADMM-Net}:  Compared with the gradient computation in  Generic-ADMM-Net, in each layer, the gradients of the parameters are the sum of  the gradients from the real and imaginary parts of the outputs.
In the Complex-ADMM-Net, the gradients of loss w.r.t. parameters can be calculated by the following formulations:
\begin{equation}
\nonumber
\frac{\partial E}{\partial \Theta} = \{\mathbf{Re}(\frac{\partial E}{\partial O_t})\mathbf{Re}(\frac{\partial O_t}{\partial \Theta_t})
+ \mathbf{Im}(\frac{\partial E}{\partial O_t})\mathbf{Im}(\frac{\partial O_t}{\partial \Theta_t})\}_{t=1}^{N_l},
\end{equation}
where $N_l$ is the  number of  net layers, $O_t$ is the output of the $t$-th layer, and $\Theta_t$ represents  the parameters in this  layer.

\section{Experiments}
In this section, we conduct  extensive experimental evaluations for the proposed ADMM-Nets.
We train and test ADMM-Nets on   brain and chest MR images~\footnote{CAF Project: \scriptsize{{https://masi.vuse.vanderbilt.edu/workshop2013/index.php\\/Segmentation-Challenge-Details}}}
datasets in experiments.
The resolution of images is $256\times256$.
For each dataset, we randomly take 100 images for training and 50 images for testing.
The ADMM-Net is separately learned for each sampling rate such as 10\%, 20\%, 30\%, 40\% and 50\%. The sampling pattern in $k$-space is the commonly used pseudo radial sampling as show in Fig. \ref{fig:v1v2}(a).
The reconstruction accuracies are reported as the average NMSE and Peak Signal-to-Noise Ratio (PSNR) over the test images. All experiments are performed on a desktop computer with Intel core i7-4790k CPU and GTX1080 GPU.
The codes to test and train the deep ADMM-Nets will be available online.


In the following, we first evaluate our approach by comparing  two versions of networks.
We then investigate the effect of network initialization on reconstruction accuracy.
We will also test the impact of different network architectures using different filter numbers (i.e., $L$),
filter sizes (i.e., $w_f$),  iteration numbers in the sub-stages (i.e., $N_t$)  and the number of stages (i.e., $N_s$).
We next compare with conventional compressive sensing MRI methods and also the state-of-the-art
methods both qualitatively and quantitatively
under different sampling rates.
Additionally, we compare with two recent deep learning approaches on the brain data.
We finally extent our method to cope with complex-valued MR image  and
noisy $k$-space data reconstruction.

\subsection{Results of Two Versions of ADMM-Nets}

We test and compare the reconstruction performance of  Basic-ADMM-Net and  Generic-ADMM-Net.
We use the brain  set  with 20\% sampling rate to train two kinds of small-scale networks respectively.
For Basic-ADMM-Net, we use the basic network settings, i.e.,
$w_f=3$, $L=8$,  $N_{s} = 4 $ and $15$.
For Generic-ADMM-Net, we use the similar network settings, i.e.,
$w_f=3$, $L=8$, $N_{t}=1$, $N_{s} = 4 $ and $15$.
\begin{table}[!htbp]
  \caption{The result comparisons of different ADMM-Nets on brain data.} 
  \label{tab:v1v2}
  \centering
\footnotesize{
  \begin{tabular}{cccccc}
    \toprule
    \multirow{2}{*}{Method}  &\multirow{2}{*}{ RecPF} & Basic &Generic &Basic &Generic             \\
                 &         & ($N_{s}=4$)          & ($N_{s}=4$)           &($N_{s}=15$)       &($N_{s}=15$)   \\

    \midrule
     NMSE       &0.0917    &0.0858                  &0.0751                  &0.0752                 &\bf{0.0670 }           \\
     PSNR       &35.32     &35.83                   &37.04                   &37.01                  &\bf{38.03}             \\
    \bottomrule
  \end{tabular}
}
\end{table}
\begin{figure}[h]
\centering
\includegraphics[width=1\linewidth]{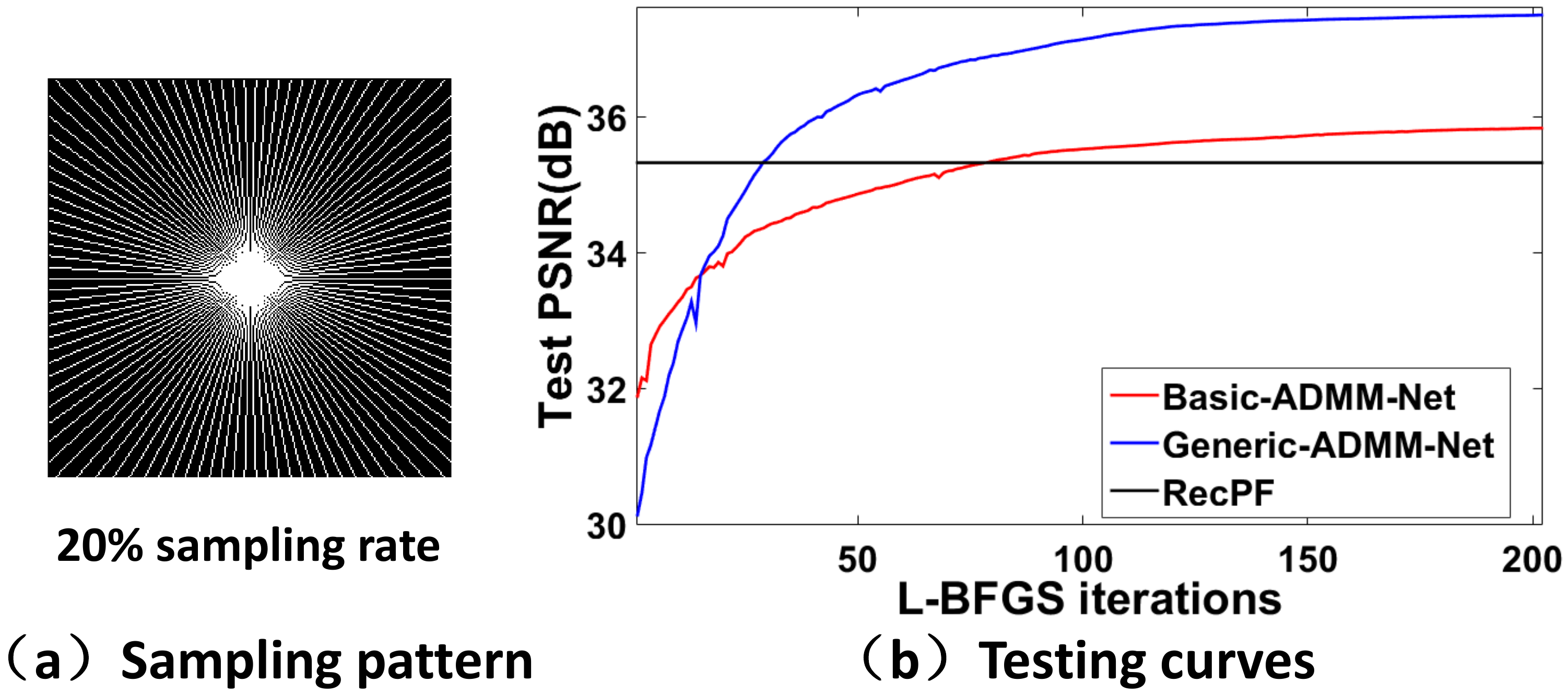}

\caption{(a) Pseudo radial sampling pattern with 20\% sampling rate.
(b) Average test PSNR comparisons between two ADMM-Nets  and baseline method RecPF with different L-BFGS iterations. }
\label{fig:v1v2}
\vspace{-0.2cm}
\end{figure}

The average PSNR curves of   Basic-ADMM-Net, Generic-ADMM-Net  and RecPF method~\cite{yang2010fast}
are shown in Fig. \ref{fig:v1v2}(b). We use the RecPF method solving a {\em TVL$_{1}$-L$_{2}$} model by ADMM algorithm as our baseline method.
As can be observed, two ADMM-Nets  can get higher performance with increase of the L-BFGS iterations.
Furthermore, the Generic-ADMM-Net can converge faster than Basic-ADMM-Net and get the highest precision.

The quantitative comparison results are shown in Tab.\ref{tab:v1v2}.
The results explicitly show that the Generic-ADMM-Net  achieves $38.03$ dB, $1.02$ dB higher than the Basic-ADMM-Net with stage $15$.
Therefore we adopt the Generic-ADMM-Net as the default network in the following experiments,
i.e., the ADMM-Net specifically refers to the Generic-ADMM-Net in the following sections.

\subsection{Effect of Different Network Initialization}
As shown in Sect.~\ref{sec:int}, the network parameters can be initialized by model-based initialization method or
random initialization method.
We train two networks with these two  initializations.
In both networks, we set
$w_f=3$, $L=8$ and $N_{t} = 1 $.
\begin{table}[!htbp]
  \caption{The result comparisons of different initializations. }
  \label{tab:int}
  \centering
\footnotesize{
  \begin{tabular}{ccccc}
    \toprule
    \multirow{2}{*}{Method}        & Model-based               &Random               &Model-based             &Random             \\
                                   & ($N_{s}=4$)       & ($N_{s}=4$)         &($N_{s}=15$)    &($N_{s}=15$)   \\

    \midrule
     NMSE                          &{0.0751}             &0.0754               &\bf{0.0670}          &0.0676           \\
     PSNR                          &{37.04}              &37.01                &\bf{38.03}           &37.97             \\
    \bottomrule
  \end{tabular}
}
\end{table}
The results in Tab.~\ref{tab:int} show that the model-based initialization method produces  marginally higher reconstruction accuracy  than random initialization with the same number of stages.
However, the model-based initialization method which uses a set of DCT bases has a limitation on the number of filters in the convolution layers, e.g., 8 filters in size of $3\times 3$ (the first DCT basis is discarded), 24 filters in size of $5\times 5$ or  48 filters in size of $7\times 7$, etc.
We will discuss the effect of the filter number on the reconstruction performance in the following.
\subsection{Performances of Different Network Architecture}

We next test  different network architectures by varying the network width (i.e., the number of filters and  filter sizes)
and network depth (i.e., the number of iterations in the sub-stages and stages).
Generally, a wider and deeper network produces higher performance, but costs more computational overhead.
We will investigate several configurations of the network structures to best trade-off between reconstruction accuracy and speed for practical demand.

\subsubsection{Network Width}

In this section, we discuss the effect of the network width on the network performance.
The width of a network is represented by the number of filters (i.e., $L$) and the size of filters (i.e., $w_f$).
First, we respectively train ADMM-Nets with $8$ filters,  $64$ filters and $128$ filters in size of $3\times 3$ in each stage.
Three nets share the same network settings  which are  $w_f =3$, $N_{s} = 4 $ and $N_{t}=1$.
\begin{table}[!htbp]
  \caption{The result comparisons of different filter numbers on brain data.}
  \label{tab:filterN}
  \centering
\footnotesize{
  \begin{tabular}{cccc}
    \toprule
    {Filter Number}      & L=8           &L=64             &L=128            \\
    \midrule
     NMSE                &0.0754         &0.0730           &{\bf0.0701}            \\
     PSNR                &37.01          &37.36            &{\bf37.64}              \\
     CPU$\backslash$GPU Time   &0.080$\backslash$0.039s      &0.451$\backslash$0.138s   &0.855$\backslash$0.258s                   \\
    \bottomrule
  \end{tabular}
}
\end{table}
The reconstruction results on brain data with 20\% sampling rate are shown in Tab.~\ref{tab:filterN}.
It clearly shows that using more filters could inherently improve the reconstruction quality.
The net with 128 filters achieves highest accuracy, e.g., it achieves 0.28 dB higher than the net with 64 filters, but takes around  two times in  the running time.
Then, we justify the effect of different filter sizes in ADMM-Net for reconstruction.
We examine three networks with different filter sizes, i.e., $w_f=3$, $w_f=5$ and $w_f=7$.
Other parameters of the network are  $L =128$, $N_{s} = 4 $ and  $N_{t}=1$.
The results in Tab.~\ref{tab:filterS} show that using  larger filter size can effectively  improve the result.
This maybe because larger filter size increases the receptive field of network output  modeling higher-order image local correlations.
However, the reconstruction speed will be declining by using larger filters.
The performance of the network with $7\times7$ filters  is only 0.08 dB higher than  the network with $5\times5$ filters,
but it  cost longer testing time.

\begin{table}[!htbp]
  \caption{The result comparisons of different filter sizes on brain data.}
  \label{tab:filterS}
  \centering
\footnotesize{
  \begin{tabular}{cccc}
    \toprule
    {Filter Size}           & f=3                   &f=5                      &f=7            \\
    \midrule
  NMSE                   &0.0701                    &0.0657                   &{\bf0.0650}          \\
  PSNR                   &37.64                     &38.21                    &{\bf38.29}            \\
  CPU$\backslash$GPU Time&0.855$\backslash$0.258s   &1.584$\backslash$0.266s  &2.646$\backslash$0.350s               \\
    \bottomrule
  \end{tabular}
}
\end{table}

\subsubsection{Network Depth}
We further investigate the effect of different network depths on the network performance.
The  depth of ADMM-Net depends on the number of iterations in the sub-stages corresponding to  the gradient descent iterations in Eqn.~(\ref{eq:iters2}), and the number of stages  corresponding to the iterations in the ADMM algorithm.
We first train  deeper networks  by increasing the  number of iterations in the sub-stages (i.e., $N_{t}$) in each stage.
We  train and test ADMM-Nets respectively with $N_{t}=1, 2, 3$. Other network parameters are set as $N_s=3$, $w_f=5$ and $L=128$.
The first four columns of the Tab.~\ref{tab:sublayer} show the testing results of  ADMM-Nets  learned from brain data with 20\% sampling rate.
\begin{table}[!htbp]
  \caption{The result comparisons of different sub-stage numbers on brain data}
  \label{tab:sublayer}
  \centering
\footnotesize{
  \begin{tabular}{ccccc}
    \toprule
    \multirow{2}{*}{Depth}    & $N_{t}$=1              &$N_{t}$=2                &$N_{t}$=3       & $N_{t}$=1            \\
                              & ($N_{s}=3$)      &($N_{s}=3$)        &($N_{s}=3$)      &($N_{s}=4$)    \\
    \midrule
     NMSE                     &0.0680            &0.0670             &0.0666           & 0.0657          \\
     PSNR                     &37.90             &38.03              &38.08            & 38.21           \\
    \bottomrule
  \end{tabular}
}
\end{table}
We find that deeper networks (i.e., having more iterations in sub-stages) with same stage number always result in better accuracy.
\begin{figure}[h]
\centering
\includegraphics[width=0.9\linewidth]{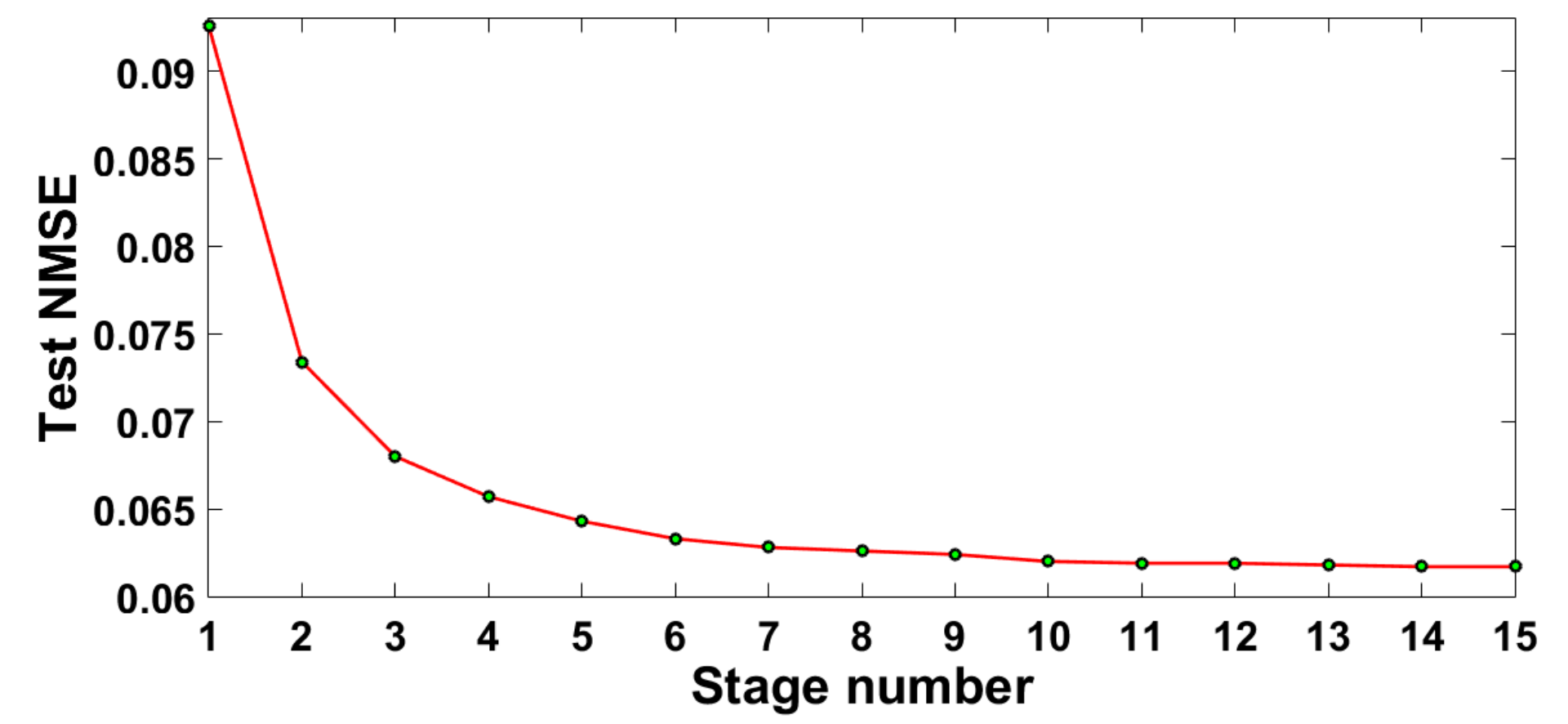}
\caption{The testing NMSEs  of ADMM-Nets training on brain data with 20\% sampling rate
using different number of stages.  }
\label{fig:stage2}
\end{figure}
\begin{table*}[!htbp]
  \caption{Performance comparisons  on brain data with different sampling rates.   }
  \label{tab:brain}
  \centering
  \begin{tabular}{llllllllllll}
    \toprule
    \multirow{2}{*}{Method}  &\multicolumn{2}{c}{10\%} &\multicolumn{2}{c}{20\%}        &\multicolumn{2}{c}{30\%}                    &\multicolumn{2}{c}{40\%}      &\multicolumn{2}{c}{50\%}     & Test Time     \\
                                \cmidrule{2-11}
                               & NMSE  & PSNR  & NMSE  & PSNR   & NMSE  & PSNR   & NMSE  & PSNR    & NMSE  & PSNR   &CPU\ $\backslash$ GPU\\
    \midrule
    Zero-filling~\cite{bernstein2001effect}                &0.2624&26.35 &0.1700 &29.96   &0.1247 &32.59   &0.0968 &34.76    &0.0770 &36.73   &0.001s$\backslash$-{}-  \\
    TV~\cite{lustig2007sparse}   &0.1539&30.90&0.0929 &35.20   &0.0673 &37.99   &0.0534 &40.00    &0.0440 &41.69   &0.739s$\backslash$-{}- \\
    RecPF~\cite{yang2010fast}    &0.1498&30.99&0.0917 &35.32   &0.0668 &38.06   &0.0533 &40.03    &0.0440 &41.71   &0.311s$\backslash$-{}-  \\
    SIDWT~\footnote{Rice Wavelet Toolbox: \scriptsize{http://dsp.rice.edu/software/rice-wavelet-\\toolbox}}                       &0.1564&30.81 &0.0885 &35.66   &0.0620 &38.72   &0.0484 &40.88    &0.0393 &42.67  &7.864s$\backslash$-{}-  \\
    \midrule
    PBDW~\cite{qu2012undersampled}&0.1290&32.45&0.0814&36.34   &0.0627 &38.64   &0.0518 &40.31    &0.0437 &41.81  &35.364s$\backslash$-{}- \\
    PANO~\cite{qu2014magnetic}  &0.1368&31.98 &0.0800 &36.52   &0.0592 &39.13   &0.0477 &41.01    &0.0390 &42.76  &53.478s$\backslash$-{}- \\
    FDLCP~\cite{zhan2015fast}    &0.1257&32.63&0.0759 &36.95   &0.0592 &39.13   &0.0500 &40.62    &0.0428 &42.00  &52.222s$\backslash$-{}- \\
    BM3D-MRI~\cite{eksioglu2016decoupled}&0.1132&33.53&0.0674 &37.98&0.0515 &40.33&0.0426 &41.99 &0.0359 &43.47  &40.911s$\backslash$-{}- \\
    \midrule
Init-Net$_{10}$           &0.2589&26.17     &0.1737&29.64   &0.1299 &32.16   &0.1025 &34.21    &0.0833 &36.01  &3.827s$\backslash$0.644s \\
ADMM-Net$_{10}$&\bf{0.1082}&\bf{33.88}&\bf{0.0620}&\bf{38.72}&\bf{0.0480}&\bf{40.95}&\bf{0.0395}&\bf{42.66}&\bf{0.0328}&\bf{44.29} &3.827s$\backslash$0.644s \\
    \bottomrule
  \end{tabular}
\end{table*}
Then to test the effect of  stage number (i.e., $N_s$), we 
train deeper networks by adding one stage at each time with other parameters fixed, i.e., $N_{t}=1$, $w_f=5$ and $L=128$.
Figure~\ref{fig:stage2} shows the average  testing NMSE values of the ADMM-Nets  using different number of stages under the 20\% sampling rate by directly joint training.
The reconstruction error decreases fast when $N_s \leq 10$ and  marginally decreases when further increasing the number of stages.
The testing time of the net increases  linearly along with the number of stages.

In the deep ADMM-Net,
an increase in the number of parameters by adding one iteration in the sub-stage is similar to an increase by adding one stage.
However, from the Tab.~\ref{tab:sublayer},
the reconstruction accuracy  of  the ADMM-Net with $N_{t}=1$ and $N_{s}=4$  (i.e., adding one stage)
is significantly higher than the ADMM-Net with $N_{t}=2$ and $N_{s}=3$ (i.e., adding one iteration in each sub-stage)
and the ADMM-Net with $N_{t}=3$ and $N_{s}=3$ (i.e., adding two iterations in each sub-stage).
In this case, adding more stages is more effective than adding more iterations in the sub-stages for achieving superior performance.
Therefore we adopt the ADMM-Net with the default network settings, i.e., $w_f=5$, $L=128$, $N_{t}=1$, $N_{s}=10$  in the following experiments considering the image reconstruction quality and speed.

\subsection{Comparison}

In this section, we compare our deep ADMM-Net to conventional compressive sensing MRI methods and the state-of-the-art methods
on brain data and  chest data with different sampling rates.
For each sampling rate (i.e., 10\%, 20\%, 30\%, 40\%, 50\%), we train a specific network for testing.
\begin{figure}[h]
\centering
\includegraphics[width=0.65\linewidth]{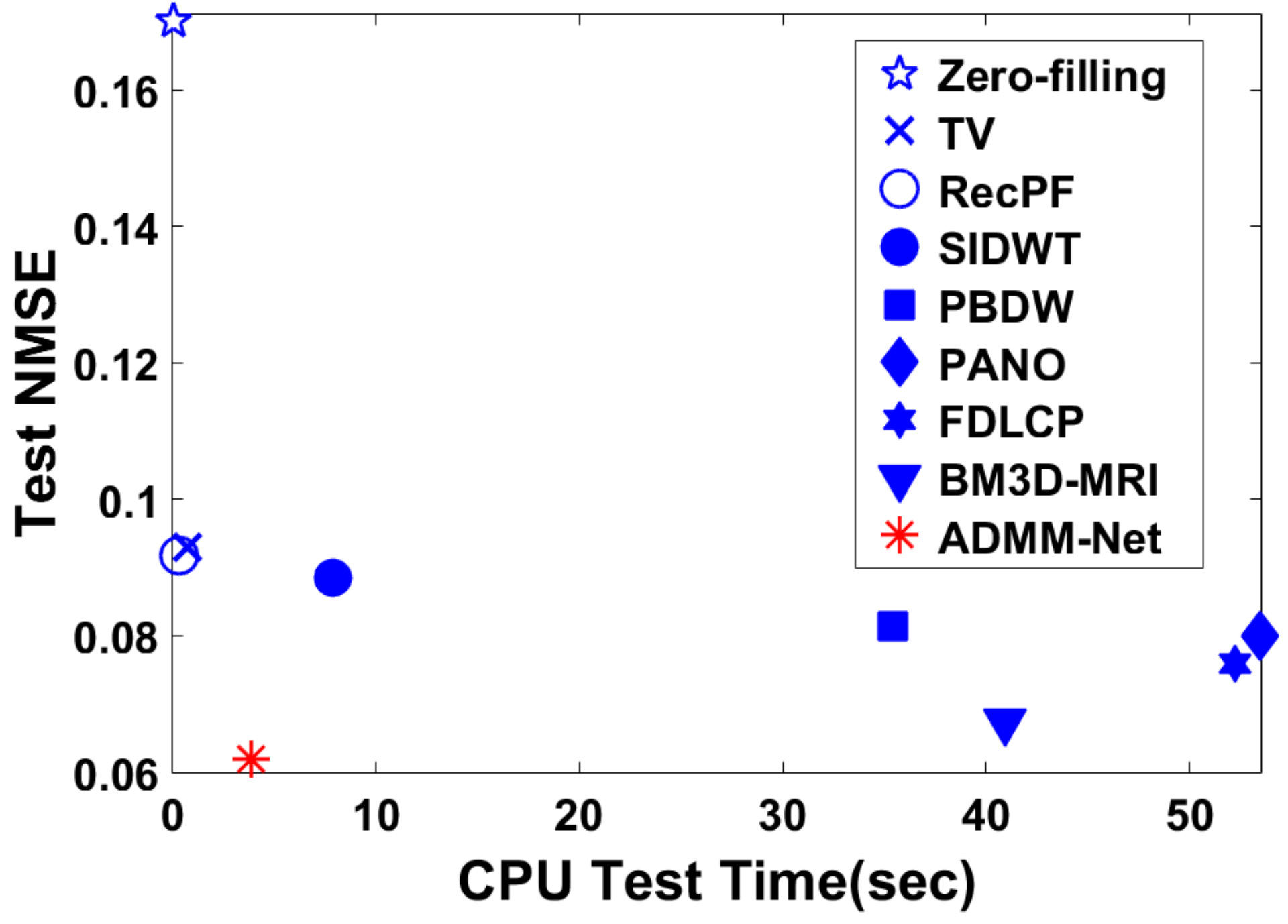}
\caption{Scatter plot of NMSEs and CPU test time for different methods. }
\label{fig:scatter}
\end{figure}
\begin{figure*}[htbp]
\centering
\includegraphics[width=0.7\linewidth]{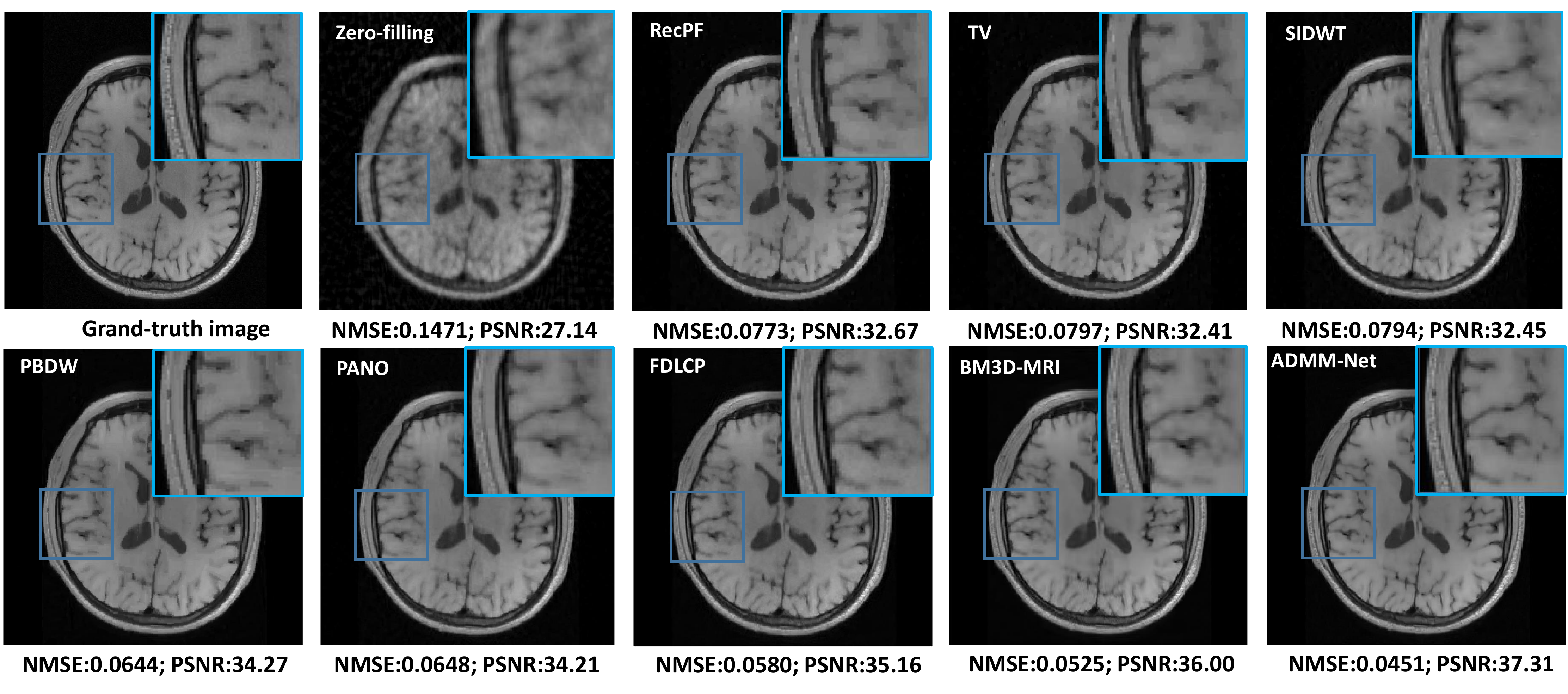}
\caption{Examples of reconstruction results on the brain data with 20\% sampling rate. }
\label{fig:20res}
\vspace{-0.2cm}
\end{figure*}
\begin{figure*}[!h]
\centering
\includegraphics[width=0.7\linewidth]{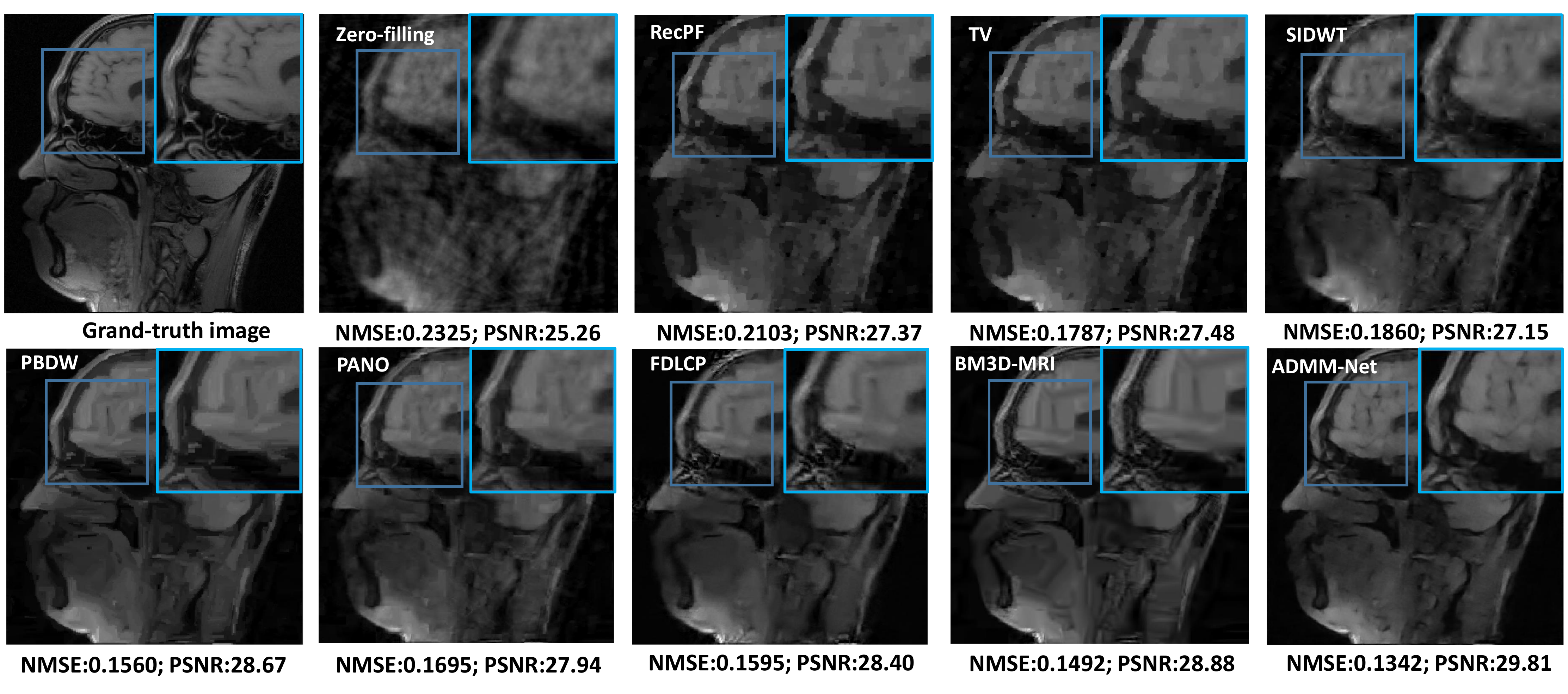}
\caption{Examples of reconstruction results on the brain data with  10\% sampling rate. }
\label{fig:10res}
\vspace{-0.2cm}
\end{figure*}

\begin{figure*}[h]
\centering
\includegraphics[width=0.9\linewidth]{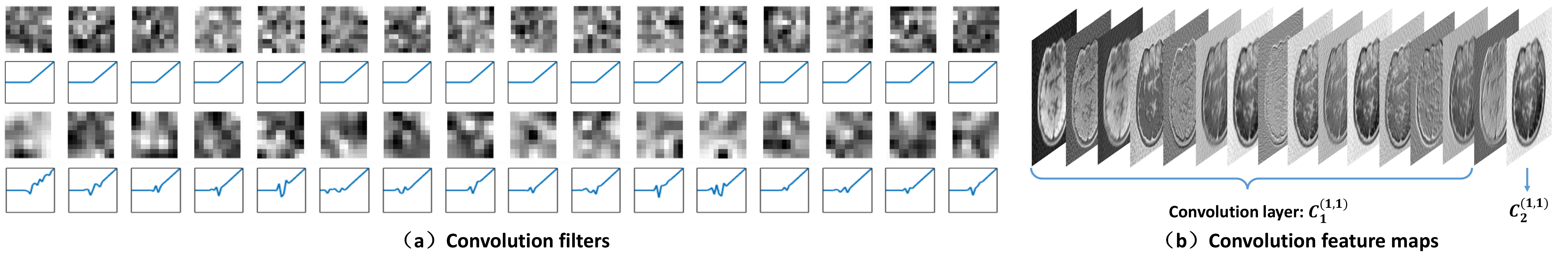}
\caption{(a) Examples of learned 16 filters  and the corresponding nonlinear transforms.
The first two rows are initial filters and nonlinear transforms, the last two rows are trained  filters and trained nonlinear transforms. (b) Example feature maps of the convolution layers in the first stage.}
\label{fig:filter}
\vspace{-0.1cm}
\end{figure*}
\begin{figure}[h]
\centering
\includegraphics[width=0.7\linewidth]{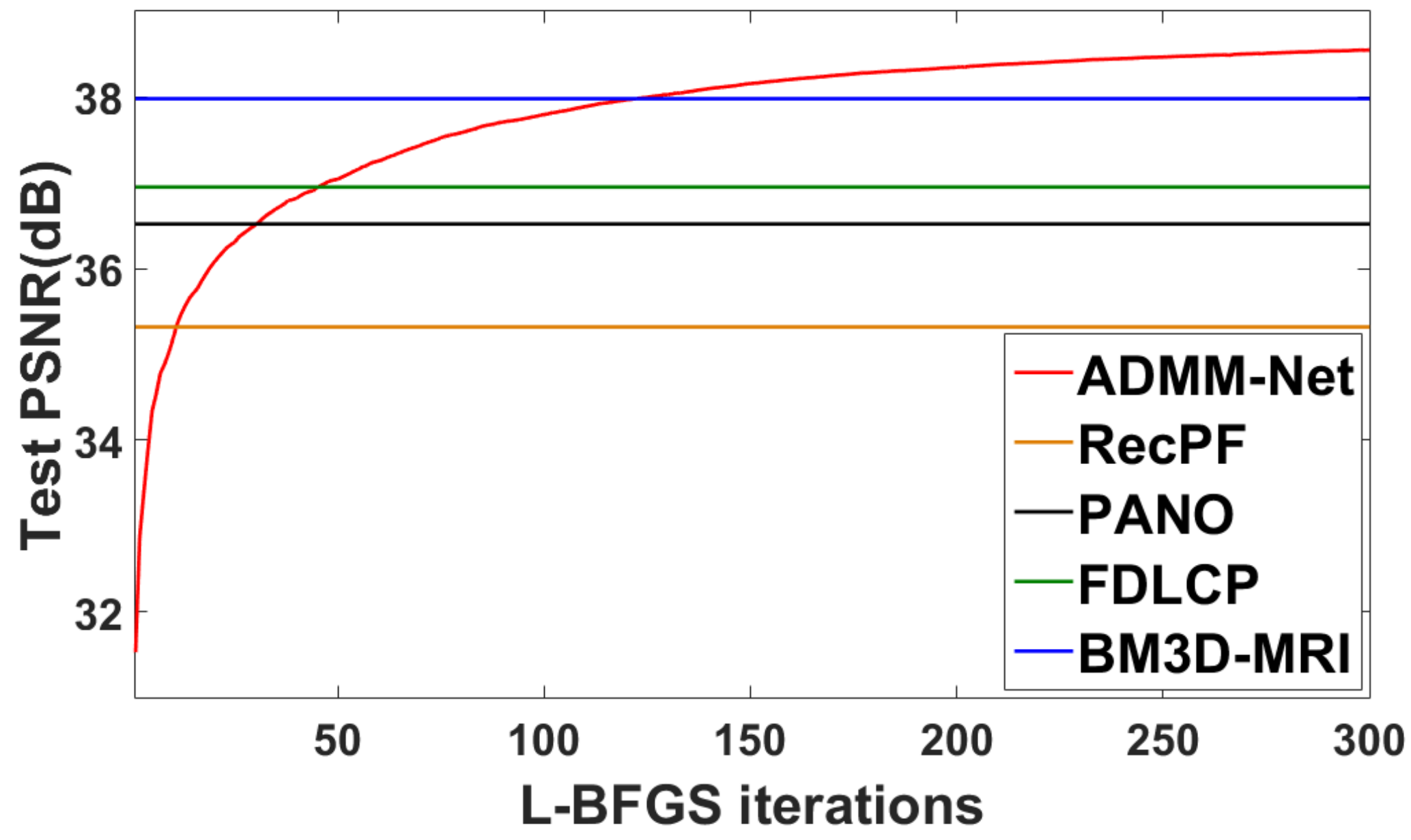}
\caption{The testing curves of different methods  with L-BFGS iterations.  }
\label{fig:able}
\vspace{-0.1cm}
\end{figure}

The conventional methods include Zero-filling~\cite{bernstein2001effect}, TV~\cite{lustig2007sparse}, RecPF~\cite{yang2010fast} and
SIDWT~\footnote{Rice Wavelet Toolbox: \scriptsize{http://dsp.rice.edu/software/rice-wavelet-\\toolbox}}.
The state-of-the-art methods include  PBDW~\cite{qu2012undersampled}  taking advantage of  patch-based directional wavelets, PANO~\cite{qu2014magnetic}  making use of similarity of image patches, 
FDLCP~\cite{zhan2015fast} which is a  variant of dictionary learning method
and BM3D-MRI~\cite{eksioglu2016decoupled}
relying on a well designed BM3D
denoiser.
For fair comparison, 
the adjustments of  all  parameters are according to the same validation set.

\begin{table*}[!h]
  \caption{Comparisons of NMSE and PSNR on chest data with 20\% sampling rate. }
  \label{tab:chest}
  \centering
\footnotesize{
  \begin{tabular}{ccccccccccc}
    \toprule
    Method  &Zero-filling &TV     &RecPF &SIDWT &PBDW      &PANO     & FDLCP    &BM3D-MRI   & ADMM-Net$_{Brain}$    &ADMM-Net$_{10}$   \\
    \midrule
     NMSE  &0.1903    &0.1019 &0.1017 &0.0952&0.0884&0.0858    &0.0775   &0.0694     &0.0759                 &\bf{0.0668} \\
     PSNR  &30.04     &35.49  &35.51  &36.15&36.69&37.01     &37.77    &38.79      &38.00                  &\bf{39.10}\\
    \bottomrule
  \end{tabular}
}
\vspace{-0.2cm}
\end{table*}
Table \ref{tab:brain} shows the quantitative results of different methods in different sampling rates on the  brain data.
Compared with the conventional methods such as Zero-filling, TV, RecPF and SIDWT, our proposed method produces the best quality with comparable reconstruction speed in all sampling rates.
Compared with the state-of-the-art methods such as PBDW, PANO, FDLCP and BM3D-MRI, the ADMM-Net still has the most accurate reconstruction results with fastest computational speed.
Specifically, for the sampling rate of 30\%,  ADMM-Net$_{10}$ (i.e. ADMM-Net with 10 stages) outperforms the state-of-the-art methods PANO and FDLCP by 1.82 dB. Moreover, our reconstruction speed is around 13 times faster.
ADMM-Net$_{10}$ also produces 0.62 dB higher accuracy and runs around 10 times faster in computational time than BM3D-MRI method.
In Fig.~\ref{fig:scatter},
we compare the NMSEs and the average CPU testing time for different methods on 20\% brain data using scatter plot. It is easy to observe that our method is the best considering  the reconstruction accuracy and running time.


The visual comparisons of  20\% and 10\% sampling rates  are in Fig.~\ref{fig:20res} and Fig.~\ref{fig:10res}.
It clearly show that the proposed network can achieve better reconstruction qualities and  preserve the fine image details without obvious artifacts.
For the case of low sampling rate (i.e., 10\% sampling rate),
the ADMM-Net can still reconstruct structural information of the MR image while other methods failed.
The examples of the learned nonlinear functions and the filters in the  first stage trained on  brain data with 20\% sampling rate
 are shown in Fig.~\ref{fig:filter}(a).
Each filter and each non-linear function
have specific different forms  from initialization after training.
Figure~\ref{fig:filter}(b) shows an example of relevant convolution feature maps in the first stage, which include different structures and luminance information.

To demonstrate the effectiveness of network training, in Tab.~\ref{tab:brain}, we also present the results of the initialized network (i.e., Init-Net$_{10}$) for ADMM-Net$_{10}$. The network after training produces significantly improved  reconstruction accuracy, e.g., PNSR is increased from 29.64 dB to 38.72 dB with sampling rate of 20\%.
The testing curves of different methods in Fig. \ref{fig:able} show that deep ADMM-Net has powerful abilities of learning, e.g.,
 ADMM-Net$_{10}$  outperforms the baseline method (i.e., RecPF) with less 20 L-BFGS iterations,
and exceeds state of art methods  (i.e., PANO, FDLCP and BM3D-MRI) with less 150 L-BFGS iterations.

We also compare the reconstruction results on chest data with sampling rate of 20\%.
We test the  generalization ability of ADMM-Net by applying the learned net from brain data to chest data.
Table~\ref{tab:chest} shows that our net learned from brain data (ADMM-Net$_{Brain}$) still achieves competitive reconstruction accuracy on chest data, resulting in remarkable a generalization ability. This might be due to that the learned filters and nonlinear transforms are performed over local patches, which are repetitive across different organs.
Moreover, the ADMM-Net$_{10}$ learned from chest data achieves the best reconstruction accuracy.
Thus, the deep ADMM-Net can achieve high performance on MRI reconstruction with different organs.
\begin{table}[!htbp]
  \caption{Performance comparisons with deep learning methods on brain data.}
  \label{tab:deep}
  \centering
\footnotesize{
  \begin{tabular}{ccccc}
    \toprule
    Method           &Wang et al.      &Lee et al.         &ADMM-Net    \\
    \midrule
     NMSE            &0.0973      &0.1607          &{\bf0.0620}      \\
     PSNR            &34.80       &30.44           &{\bf38.72}       \\
     Training Time   &120h        &11h             &10h         \\
    \bottomrule
  \end{tabular}
}
\vspace{-0.2cm}
\end{table}

\begin{table*}[!htbp]
  \caption{Comparisons of NMSE and PSNR on complex-valued brain data with different noise level (20\% sampling rate). }
  \label{tab:complex}
  \centering
\footnotesize{
  \begin{tabular}{cccccccccccc}
    \toprule
\multicolumn{2}{c}{Method}               &Zero-filling &TV    &RecPF &SIDWT &PBDW  &PANO   &FDLCP    &BM3D-MRI  &Complex-ADMM-Net$_{10}$   \\
    \midrule
  \multicolumn{2}{c}{NMSE}   &0.2250       &0.1336&0.1320&0.1328&0.1277&0.1300   & 0.1252  &-{}-      &\bf{0.0988} \\
  \multicolumn{2}{c}{PSNR}            &28.25        &32.49 &32.54 &32.60 &32.78 &32.88    & 33.44   &-{}-      &\bf{35.40}\\
    \midrule
\multirow{2}{*}{$ \sigma = 0.010$}&NMSE  &0.2250       &0.1426&0.1427&0.1415&0.1360&0.1364  &0.1354   &-{}-      &\bf{0.0988}\\
                                  &PSNR  &28.25        &32.06 &31.94 &32.14 &32.30 &32.54  &32.81    &-{}-      &\bf{35.37}\\
    \midrule
\multirow{2}{*}{$ \sigma = 0.015$}&NMSE  &0.2250       &0.1526&0.1534&0.1500&0.1429&0.1365
&0.1463   &-{}-      &\bf{0.0989}\\
                                  &PSNR  &28.25        &31.54 &31.38 &31.72 &31.99 &32.53 &32.20    &-{}-      &\bf{35.37}\\
    \bottomrule
  \end{tabular}
}

\end{table*}

In addition, we compare to several recently proposed  deep learning approaches for compressive sensing MRI on brain data with 20\% sampling rate.
Wang et al.~\cite{Wang2016Accelerating} trained a deep convolution network to learn a mapping from down-sampled reconstruction images to  fully sampled reconstruction. Then, the output of this network is used as
the regularization term in a classical CS-MRI model (i.e. Zero-filling method).
Lee et al.~\cite{unet} adopt  U-net to accelerate the data
acquisition process, which is a deep deconvolution network with contracting path to learn the residual of MRI reconstruction.
We train these two networks on our 100 brain training images which are used for ADMM-Net.
We implement the method from Wang et al. based on the {\em caffe} code from~\cite{Dong2016Image}, and
the U-net from Lee et al. according to the {\em keras} code~\footnote{retina-unet: \scriptsize{https://github.com/orobix/retina-unet}}.
We test the networks on the same 50 brain data with sampling rate of 20\%.
From the testing results in Tab. \ref{tab:deep}, we observe that ADMM-Net outperforms these deep learning methods not only for its
reconstruction quality but also in terms of the training time.
Therefore  our method is superior among these deep learning methods, which include convolution neural network structures.


\subsection{Complex-valued and Noisy Data Reconstruction}
To evaluate the reconstruction performance of the ADMM-Net to cope with complex-valued MR
image, we train the Complex-ADMM-Net on the 100 complex-valued brain training data with the sampling rate of 20\%.
The first two lines in Tab. \ref{tab:complex}  shows the complex-valued image reconstruction results, where BM3D-MRI method  can not reconstruct a complex-valued MR image directly.
It is clear that our Complex-ADMM-Net remarkably outperforms all the CS-MRI methods.
For the sampling rate of 20\%, our method (Complex-ADMM-Net$_{10}$) outperforms
the state-of-the-art method FDLCP by 1.96 dB.
This might because of the fact that the real and imaginary parts of the data
help each other for reconstruction by sharing the network parameters.


Finally, we extent our method to reconstruct k-space data with noise for demonstrating the ability of the ADMM-Net in handling noise.
We add the Gaussian white noise to both real and imaginary parts of the training k-space data and testing k-space data.
The noisy levels are set into the range of [0, 0.02] for training data,
and the noisy levels are set to be 0.010 and 0.015 for testing.
We train a Complex-ADMM-Net using noisy training brain data with 20\% sampling rate.
Table \ref{tab:complex} shows the compare reconstruction results of different methods in different noise levels.
The hyper-parameters in compared methods are adjusted  by noisy validation data whose noisy levels are same with testing data.
As a result, the reconstruction performances of the ADMM-Net are mildly influenced by additive noise, e.g.,
the PSNR is reduced from 35.40 dB to 35.37 dB by adding noise with noise level 0.010 and 0.015.
However, the PSNR of state-of-the-art method FDLCP  has decreased  1.24 dB since the noise level is 0.015.
In summary, the deep ADMM-Net outperforms other CS-MRI methods in term of better reconstruction precision and
robustness for noise.

\section{conclusion}
In this work, we proposed  two  novel deep networks, i.e., Basic-ADMM-Net and Generic-ADMM-Net, for fast compressive sensing MRI. These two  nets are deep architectures defined over  data flow graphs determined by  ADMM algorithms optimizing a general CS-MRI model. This proposed method naturally combines the advantages of model-based approach in easily incorporating domain knowledge, and the deep learning approach in effective parameter learning. Extensive experiments show that these deep nets achieved higher reconstruction accuracy while keeping the computational efficiency of the ADMM algorithm.
As a general framework, the idea that reformulates an ADMM algorithm as a deep network can also be  applied to other applications
such as image deconvolution, super-resolution, etc.

%

%

\ifCLASSOPTIONcompsoc
  \section*{Acknowledgments}
\else
  \section*{Acknowledgment}
\fi
The authors would like to thank Prof. Dong Liang
at Shenzhen Institutes of Advanced Technology for providing the brain MRI data used
in the experiments.
\ifCLASSOPTIONcaptionsoff
  \newpage
\fi



%
\bibliographystyle{ieeetr}
\bibliography{Net1}
\end{document}